\documentclass[sigconf]{acmart}
\usepackage{subcaption}
\usepackage{pifont}
\usepackage{tabularx}
\usepackage{enumitem}
\usepackage{amsthm}
\usepackage{microtype}
\usepackage[american]{babel}
\usepackage{xcolor}
\usepackage[ruled]{algorithm2e}

\SetCommentSty{mycommfont}

\SetKwInput{KwInput}{Input}                
\SetKwInput{KwOutput}{Output}              

\newcommand{\squishlist}{
 \begin{list}{$\bullet$}
  { \setlength{\itemsep}{0pt}
     \setlength{\parsep}{3pt}
     \setlength{\topsep}{3pt}
     \setlength{\partopsep}{0pt}
     \setlength{\leftmargin}{1.5em}
     \setlength{\labelwidth}{1em}
     \setlength{\labelsep}{0.5em} } }

\newcommand{\squishlisttwo}{
 \begin{list}{$\bullet$}
  { \setlength{\itemsep}{0pt}
     \setlength{\parsep}{0pt}
    \setlength{\topsep}{0pt}
    \setlength{\partopsep}{0pt}
    \setlength{\leftmargin}{2em}
    \setlength{\labelwidth}{1.5em}
    \setlength{\labelsep}{0.5em} } }

\newcommand{\squishend}{
  \end{list}  }

\AtBeginDocument{%
  \providecommand\BibTeX{{%
    \normalfont B\kern-0.5em{\scshape i\kern-0.25em b}\kern-0.8em\TeX}}}

\setcopyright{acmcopyright}
\copyrightyear{2018}
\acmYear{2018}
\acmDOI{10.1145/1122445.1122456}

\acmConference[Woodstock '18]{Woodstock '18: ACM Symposium on Neural
  Gaze Detection}{June 03--05, 2018}{Woodstock, NY}
\acmBooktitle{Woodstock '18: ACM Symposium on Neural Gaze Detection,
  June 03--05, 2018, Woodstock, NY}
\acmPrice{15.00}
\acmISBN{978-1-4503-XXXX-X/18/06}

\def\b{{\bf b}}

\def\e{{\bf e}}

\def\v{{\bf v}}

\def\W{{\bf W}}

\def\0{{\bf 0}}
\def\1{{\bf 1}}
\def\2{{\bf 2}}
\def\3{{\bf 3}}
\def\4{{\bf 4}}
\def\5{{\bf 5}}
\def\6{{\bf 6}}
\def\7{{\bf 7}}
\def\8{{\bf 8}}
\def\9{{\bf 9}}

\copyrightyear{2021}
\acmYear{2021}
\setcopyright{rightsretained}
\acmConference[KDD '21]{Proceedings of the 27th ACM SIGKDD Conference on Knowledge Discovery and Data Mining}{August 14--18, 2021}{Virtual Event, Singapore}
\acmBooktitle{Proceedings of the 27th ACM SIGKDD Conference on Knowledge Discovery and Data Mining (KDD '21), August 14--18, 2021, Virtual Event, Singapore}\acmDOI{10.1145/3447548.3467304}
\acmISBN{978-1-4503-8332-5/21/08}

\begin{document}
\fancyhead{}

\title{Learning to Embed Categorical Features without Embedding Tables for Recommendation}


\author{Wang-Cheng Kang, Derek Zhiyuan Cheng, Tiansheng Yao, Xinyang Yi, Ting Chen, Lichan Hong, Ed H. Chi}
\email{{wckang, zcheng, tyao, xinyang, iamtingchen, lichan, edchi}@google.com}
\affiliation{%
  \institution{Google Research, Brain Team}
}
\renewcommand{\shortauthors}{W.-C. Kang, et al.}

\begin{abstract}
Embedding learning of categorical features (e.g. user/item IDs) is at the core of various recommendation models including matrix factorization and neural collaborative filtering. The standard approach creates an embedding table where each row represents a dedicated embedding vector for every unique feature value. However, this method fails to efficiently handle high-cardinality features and unseen feature values (e.g. new video ID) that are prevalent in real-world recommendation systems.
In this paper, we propose an alternative embedding framework Deep Hash Embedding (DHE), replacing embedding tables by a deep embedding network to compute embeddings on the fly.
DHE first encodes the feature value to a unique identifier vector with multiple hashing functions and transformations, and then applies a DNN to convert the identifier vector to an embedding.
The encoding module is deterministic, non-learnable, and free of storage, while the embedding network is updated during the training time to learn embedding generation.
Empirical results show that DHE achieves comparable AUC against the standard one-hot full embedding, with smaller model sizes.
Our work sheds light on the design of DNN-based alternative embedding schemes for categorical features without using embedding table lookup.
\end{abstract}
\vspace{-0.2cm}




\maketitle

\begin{figure}[t]
\centering
\includegraphics[width=1.00\linewidth]{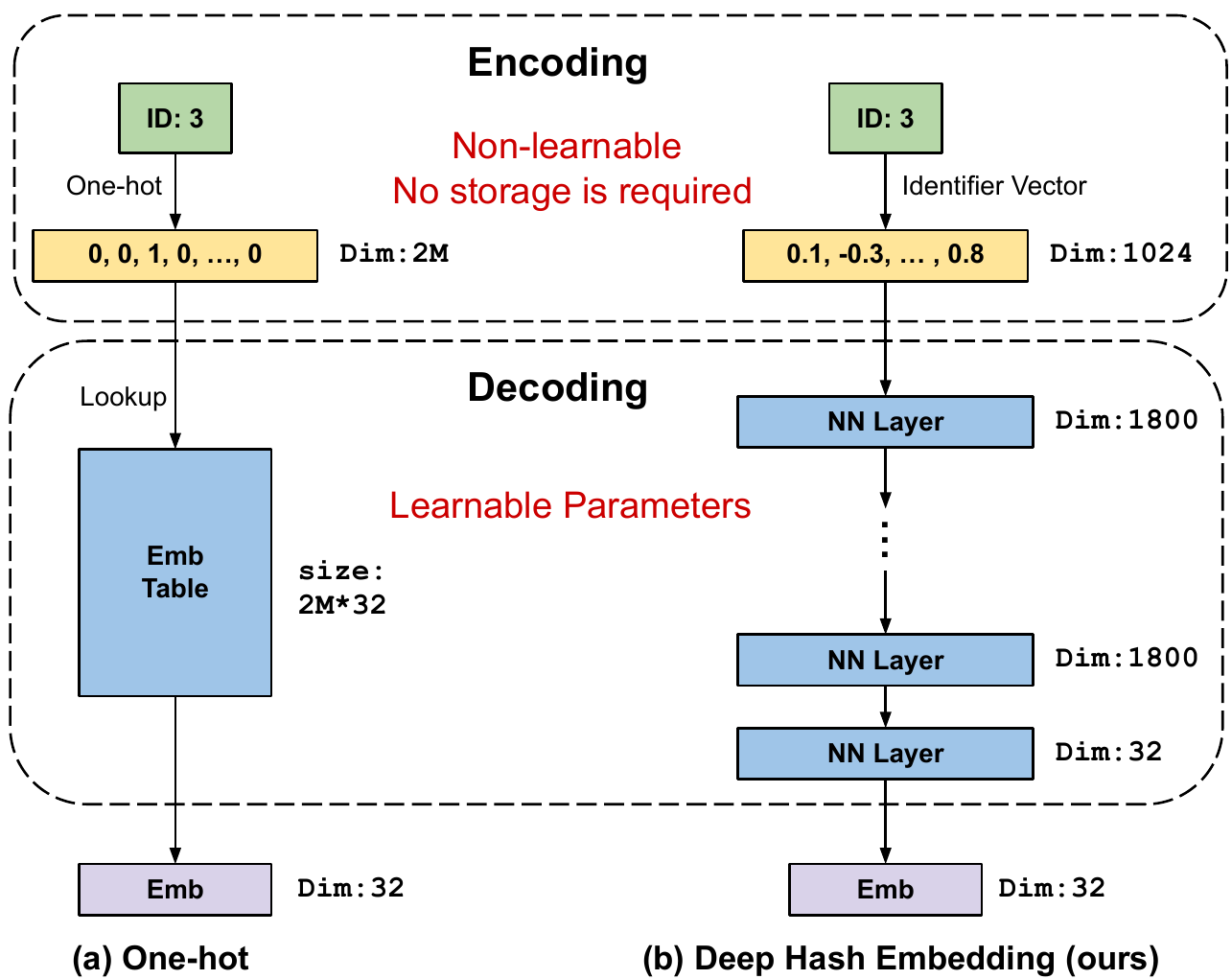}
\vspace{-0.4cm}
\caption{An illustration of one-hot based full embedding and Deep Hash Embedding~(DHE) for generating 32-dim embeddings for 2M IDs. The dimension numbers are from our experiments for providing a concrete example. The two models achieve similar AUC while DHE costs 1/4 of the full model size. DHE uses a dense hash encoding to obtain a unique identifier for each feature value, and applies a deep embedding network to generate the feature embedding. DHE doesn't perform any embedding lookup.}\label{fig:intro}
\vspace{-0.2cm}
\end{figure}

\section{Introduction}

Machine learning is highly versatile to model various data types, including continuous features, sparse features, and sequential features. Among these, we focus on improving embedding learning for large-vocabulary categorical features. Specifically, we assume a categorical feature is defined by a vocabulary $V$, with the feature value is (exactly) one of the elements in $V$. For example, ID features are typically categorical features where each \emph{feature value} is a unique ID (e.g. video ID). Another example is the ``device'' feature, and ``iPhone~12'' is a possible \emph{feature value}. 

Embedding learning has become the core technique for modeling categorical features, and have been adopted in various models, such as Matrix Factorization~(MF)~\cite{DBLP:conf/uai/RendleFGS09} and word2vec~\cite{word2vec}. The embedding learning technique greatly helps us understand the semantic meaning of feature values (e.g. words). Embeddings have also become the cornerstone of deep models for capturing more complex interactions among feature values (e.g. BERT~\cite{bert}, DeepFM~\cite{DBLP:conf/ijcai/GuoTYLH17}).

Despite the success of embedding learning in various domains like natural language processing~(NLP)~\cite{word2vec}, there are several challenges when applying embedding learning in recommendation:
\squishlist
\item \textbf{Huge vocabulary size}: Recommender systems usually need to handle high-cardinality categorical features (e.g. billions of video IDs for online video sharing platforms). Moreover, in NLP tasks, the vocabulary size of words is typically small (e.g. the advanced model BERT~\cite{bert} has a vocabulary of only 30K tokens) due to the common use of sub-word segmentation~\cite{DBLP:conf/acl/SennrichHB16a} for reducing the vocabulary size. But it's generally infeasible to apply this approach to the categorical features in recommendation.
\item \textbf{Dynamic nature of input}: Unlike vocabularies of words that are relatively static, the vocabulary in recommender systems could be highly dynamic. New users and new items enter the system on a daily basis, and stale items are gradually vanishing.
\item \textbf{Highly-skewed data distribution}: The categorical features in recommendation data usually follow highly skewed power-law distributions. The small number of training examples on infrequent feature values hurts the embedding quality for the tail items significantly.
\squishend

The one-hot encoding is widely adopted for embedding learning, that maps a feature value to a one-hot vector, and then looks up the embedding vector in an embedding table. However, the one-hot representation often results in a huge embedding table especially for a large-vocab feature, and it also fails to adapt to out-of-vocab feature values. In web-scale neural recommenders, it is not surprising to have most of the parameters spent on the embedding table, while the neural network itself only accounts for a very small portion of parameters~\cite{DBLP:conf/recsys/CovingtonAS16}. In practice, to better handle new (i.e., out-of-vocab / unseen) feature values and reduce the storage cost, the hashing trick~\cite{hashing} is often adopted, that randomly maps feature values to a smaller number of hashing buckets, though the inevitable embedding collisions generally hurt performance. Essentially, these embedding approaches can be viewed as a 1-layer \emph{wide} neural network (i.e., the embedding table) with \emph{one-hot} encoding.

In this paper, we seek to explore a \emph{deep}, \emph{narrow}, and \emph{collision-free} embedding scheme without using embedding tables. We propose the Deep Hash Embedding (DHE) approach, that uses \emph{dense} encodings and a \emph{deep} embedding network to compute embeddings on the fly. This completely replaces the traditional embedding tables. Specifically, we use multiple hashing and appropriate transformations to generate a unique, deterministic, dense, and real-valued vector as the identifier encoding of the given feature value, and then the deep embedding network transforms the encoding to the final feature embeddings. The feature embeddings are then fed into recommendation models (e.g. MF or deep recommendation models) for end-to-end training. Figure~\ref{fig:intro} depicts the comparison between the standard one-hot based embedding and DHE.
Our main contributions are listed as follows:
\begin{itemize}[leftmargin=5mm]
\item We analyze various embedding methods, including hashing-based approaches for categorical features. Unlike existing methods that heavily rely on one-hot encodings, we encode each feature value to a unique dense encoding vector with multiple hashing, which takes the first step to completely remove the huge embedding tables for large-vocab features.
\item With the dense encoding, we replace the commonly used embedding lookup (essentially a shallow and  wide network) with \emph{deep} embedding networks, which is more parameter-efficient. We also address the trainability and expressiveness issues to improve the ability of embedding generation.
\item We propose Deep Hash Embedding~(DHE) based on aforementioned encodings and deep embedding networks. We further improve DHE to better generalize among feature values and to new values, by integrating side features in the encodings. 
\item We conduct extensive experiments on two benchmark datasets for recommendation tasks with large-vocab categorical features. We compare with state-of-the-art models and analyze the effect of various key components in DHE. The results suggest that DHE is a promising alternative to one-hot full embeddings.
\end{itemize}


We first discuss various existing one-hot based embedding methods from the perspective of neural networks. Then we introduce DHE's dense hash encodings, deep embedding network, and an extension of using side features for better encodings, before we present our experimental results. Finally, we discuss related work, conclude our paper, and point out promising directions for future work.


\begin{table*}
\caption{Comparison of embedding schemes. The model size of DHE is independent of $n$ or $m$. DHE is based on dense hash encodings and deep neural networks. DHE can handle out-of-vocab values for online learning, and incorporate side features.}\vspace{-0.3cm}
\footnotesize
\begin{tabular}{lcccccc}
\toprule
                     & \textbf{Full Emb} & \textbf{The Hashing Trick~\cite{hashing}} & \textbf{Bloom Emb~\cite{bloom}} & \textbf{Compositional Emb~\cite{fb}} & \textbf{Hash Emb~\cite{hashemb}} & \textbf{Deep Hash Emb (DHE)} \\
\midrule                     
\textbf{Model Size} &   $O(nd)$      &   $O(md)$                &  $O(md)$         &     $O(md+\frac{n}{m}d^2)$           &   $O(nk\text{+}md)$       &  $O(kd_{\text{NN}}\text{+}(h\text{-}1)d_{\text{NN}}^2\text{+}dd_{\text{NN}})$   \\
\textbf{\#Hash Functions}               &   -      &    1     &     2$\sim$4     &      2          &    2      & $\sim$1000    \\
\textbf{Encoding Vector}               &    one-hot      &    one-hot     &     (multi) one-hot     &      (multi) one-hot &    (multi) one-hot & dense \& real-valued    \\
\textbf{Decoding Function}               &    1-layer NN      &    1-layer NN     &     1-layer NN     &      3-layer NN &    1-layer NN & Deep NN\\
\textbf{Emb Table Lookup?}           &\ding{52}&\ding{52}&\ding{52}&\ding{52}&\ding{52}&\ding{56}    \\
\textbf{Handling OOV Values?}      & \ding{56}  &\ding{52}&\ding{52}
&\ding{56}&\ding{52}&\ding{52}\\
\textbf{Side Features for Encoding?}        &  \ding{56}    &  \ding{56}          &    \ding{56}    &\ding{56}           &   \ding{56}    &     \ding{52}

\\

\bottomrule
\end{tabular}
\vspace{-0.3cm}
\end{table*}


\section{Preliminary: One-hot based Embedding Learning}

The core idea of embedding learning is to map feature values into a $d$-dimensional continuous space. These learnable embeddings could be utilized by shallow models like word2vec~\cite{word2vec} or MF~\cite{DBLP:conf/uai/RendleFGS09}, to directly measure the similarity between two feature values (e.g. large inner products between similar words' embeddings). Moreover, deep models like DeepFM~\cite{DBLP:conf/ijcai/GuoTYLH17} or BERT~\cite{bert}, can model more complex structures via considering interactions among the embeddings.

We define a general framework for describing various existing embedding methods as well as our proposed approach. The embedding function $\mathcal{T}: V \to R^d$ maps a feature value (from  vocabulary $V$ with size $|V|=n$) to an embedding vector $\e\in\mathbb{R}^d$. Generally, the embedding function can be decomposed into two components: $\mathcal{T}=F\circ E$, where $E$ is an encoding function to represent feature values in some spaces, and $F$ is a decoding function to generate the embedding $\v$. In this section, we introduce full and hashing-based embedding schemes with one-hot encodings. The notation is summarized in Table~\ref{tb:notation}.

\subsection{One-hot Full Embedding}

This is the most straightforward and commonly used approach to embed categorical features, which assigns each feature value a unique $d$-dimensional embedding in an embedding table. Specifically, the encoding function $E$ maps a feature value into a unique one-hot vector. In offline settings, this is easy to achieve even if the feature values are non-numeric types like string (e.g. feature values `Japan' or `India' for the categorical feature `Country'), as we can scan and obtain a one-to-one mapping from feature values to $\{1,2,\dots, n\}$.

So we assume the feature values are already mapped to $\{1,2,\dots, n\}$, then the embedding approach creates an embedding table $\W\in\mathbb{R}^{n\times d}$, and looks up its $s$-th row $\W_s$ for the feature value~$s$. This is equivalent to the following: (i) we apply the encoding function $E$ to encode feature value $s$ with a one-hot encoding vector: $E(s)$=$\b \in \{0, 1\}^{n}$ where $b_s$=$1$ and $b_j$=$0$ ($j\neq s$); (ii) we then apply the decoding function $F$, a learnable linear transformation $\W\in \mathbb{R}^{n\times d}$ to generate the embedding $\e$, that is, $\e=F(\b) = \W^T\b$. In short, the embedding lookup process can be viewed as a 1-layer neural network (without bias terms) based on the one-hot encoding.

\subsection{One-hot Hash Embedding} 

Despite the simplicity and effectiveness of full embeddings, such a scheme has two major issues in large-scale or dynamic settings: (i) the size of the embedding table grows linearly with the vocabulary size, which could cause a huge memory consumption. For example, 100-dimensional embeddings for 1 billion video IDs alone costs near 400 GB of memory; (ii) in online learning settings where new values constantly arise, the full embedding scheme fails to handle unseen (out-of-vocab) feature values. To address the above issues, various hashing-based methods have been proposed (e.g. \cite{hashing, hashemb, bloom}), and widely used in production-scale systems for handling large-vocab and out-of-vocab categorical features (e.g. \emph{Youtube}~\cite{DBLP:conf/recsys/YiYHCHKZWC19} and \emph{Twitter}~\cite{twitter}).

The hashing trick~\cite{hashing} is a representative hashing method for reducing the dimension of the one-hot encoding for large vocabularies. The encoding function $E$ still maps a feature value into a one-hot vector, but with a different (typically smaller) cardinality of $m$: $E(s)=\b \in \{0, 1\}^{m}$ where $s\in V$, $b_{H(s)}$=$1$ and $b_j$=$0\ (j\neq H(s))$. The hash function $H$ maps feature values (including unseen values) to $\{1,2,\dots,m\}$ where $m$ is the hashed vocabulary size. The hash function $H$ seeks to distribute hashing values as uniformly as possible to reduce collision, though it's inevitable when $m<n$. Similarly, the decoding function returns the $H(s)$-th row of the embedding table. In summary, the hashing trick uses hashing to map feature values into $m$-dim one-hot vectors, and then applies a 1-layer network to generate the embeddings.

Although the hashing trick is able to arbitrarily reduce the cardinality of the original vocabulary $V$, it suffers from the embedding collision problem. Even in the ideal case (uniformly distributed), each embedding (in the embedding table) is shared by $[n/m]$ feature values on average. This inevitably hurts the model performance, as the model cannot distinguish different feature values due to the same embedding representations. To alleviate this issue, multiple hash functions have been used to generate multiple one-hot encodings: $E(s)=\b=[\b^{(1)};\b^{(2)};\dots;\b^{(k)}] \in \{0, 1\}^{m*k}$ where $b^{(i)}_{H^{(i)}(s)}$=$1$ and $b^{(i)}_j$=$0\ (j\neq H^{(i)}(s))$. Here, $k$ hash functions $\{H^{(1)},H^{(2)},\dots,H^{(k)}\}$ are adopted to generate $k$ one-hot encodings $\{\b^{(1)}, \b^{(2)}, \dots, \b^{(k)}\}$, and the concatenation is used as the encoding.

The core idea is that the concatenated encodings are less likely to be collided. We can lookup $k$ embeddings in $k$ embedding tables (respectively) and aggregate them into the final embedding. A common aggregation approach is `add'~\cite{hashemb,bloom,twitter}, which can be simply expressed as $\e=F(\b)=\W^T\b=\W^T[\b^{(1)};\b^{(2)};\dots;\b^{(k)}]$. That is to say, multiple one-hot vectors are generated with different hash functions, and then the concatenation is fed into a 1-layer neural network without bias terms. It's also common to just create and share a single embedding table~\cite{bloom, hashemb, twitter}. Mathematically, it's equivalent to $\v=\W^T\b=\W^T(\b^{(1)}+\b^{(2)}+\dots+\b^{(k)})$. 
Note that existing methods didn't scale to large $k$ and the most commonly used variant is double hashing ($k$=2)~\cite{bloom, hashemb, twitter}. 

\section{Deep Hash Embeddings (DHE)}

As introduced in the previous section, both full embeddings and hashing-based embeddings methods are essentially based on one-hot encodings and shallow networks. In this section, we propose Deep Hash Embeddings~(DHE), an alternative scheme for embedding learning in large-vocab or dynamic settings. DHE uses real-valued dense encodings and deep neural networks for generating embeddings without any embedding lookup.

Following the embedding framework of encoding and decoding ($\mathcal{T}$=$F\circ E$), we propose several properties for designing good encodings, and then introduce our encoding function $E$ and the decoding function $F$ in DHE, followed by side-feature-enhanced encoding design for enabling generalization.

\subsection{Encoding Design}\label{sec:encoding}
What is a good encoding if we have no prior knowledge about feature values? This is the core question we seek to investigate in this section, and it also leads to our design of the encoding for DHE. We conclude the following properties for designing good encodings:
\begin{itemize}[leftmargin=5mm]
\item\textbf{Uniqueness}: The encoding should be unique for each feature value. This is also the target of full embedding and multiple hashing methods. Otherwise, there are feature values that have to share the same encoding. The collided encodings make the subsequent decoding function impossible to distinguish different feature values, which typically hurts model performance.
\item\textbf{Equal Similarity}: We think only having the uniqueness is not enough. An example is \textbf{binary encoding}, which uses the binary representation as the encoding of integers (e.g. IDs): e.g. $H(9)=[1,0,0,1]$. We can see that $H(8)=[1,0,0,0]$ is more similar to $H(9)$, compared with $H(7)=[0,1,1,1]$. We believe this introduces a wrong inductive bias (ID 8 and ID 9 are more similar), which may mislead the subsequent decoding function. The double hashing has a similar issue: the encodings of two feature values that collide in one hash function, are more similar than those of two values that have no collision in both hash functions. As we don't know the semantic similarity among categorical features, we should make any two encodings be equally similar, and not introduce any inductive bias.
\item\textbf{High dimensionality}: We hope the encodings are easy for the subsequent decoding function to distinguish different feature values. As high-dimensional spaces are often considered to be more separable (e.g. kernel methods), we believe the encoding dimension should be relatively high as well. For example, one-hot encoding has an extremely large dimensionality ($n$ for full embedding and $m$ for hash embedding). Another example is \textbf{identity encoding} which directly returns the ID number: e.g. $E(7)=[7]$. Although this gives a unique encoding for each ID, it'd be extremely difficult for the following decoding function to generate embeddings based on the 1-dim encoding.
\item\textbf{High Shannon Entropy}: The Shannon entropy~\cite{DBLP:journals/bstj/Shannon48} measures (in the unit of `bits') the information carried in a dimension. The high entropy requirement is to prevent redundant dimensions from the information theory perspective. For example, an encoding scheme may satisfy the above three properties, but, on some dimensions, the encoding values are the same for all the feature values. So we hope all dimensions are effectively used via maximizing the entropy on each dimension. For example, one-hot encodings have a very low entropy on every dimension, as the encoding on any dimension is 0 for most feature values. Therefore, one-hot encodings need extremely high dimensions (i.e., $n$) and are highly inefficient.
\end{itemize}

The formal definitions and analysis of the encoding properties can be found in Appendix, and we summarize the results in Table~\ref{tab:encod}.

\begin{table}[t]
\centering
\caption{Encoding comparison regarding the four properties: U: uniqueness; E-S: equal similarity; H-D: high-dimensionality; H-E: high entropy. \label{tab:encod}}\vspace{-0.3cm}
\small\setlength{\tabcolsep}{7pt}
\begin{tabular}{lccccc}
\toprule
 Encoding	&Length&  U  &  E-S  & H-D & H-E\\ \midrule
\textbf{One-hot}&$n$&\ding{52}&\ding{52}&\ding{52}&\ding{56}\\
\textbf{One-hot Hash}&$m$&\ding{56}&\ding{52}&\ding{52}&\ding{56}\\
\textbf{Double One-hot Hash}&$2m$&\ding{56}&\ding{52}&\ding{52}&\ding{56}\\
\textbf{Binary }&$\left\lceil\log n\right\rceil$&\ding{52}&\ding{56}&\ding{56}&\ding{52}\\
\textbf{Identity}&$1$&\ding{52}&\ding{56}&\ding{56}&\ding{52}\\
\textbf{DHE (Dense Hash)}&$k$&\ding{52}&\ding{52}&\ding{52}&\ding{52}\\

\bottomrule
\end{tabular}
\vspace{-0.4cm}
\end{table}

\subsection{Dense Hash Encoding}

After analyzing the properties of various encoding schemes, we found no existing scheme satisfies all the desired properties. Especially we found non-one-hot based encodings like binary and identity encodings are free of embedding tables, but fail to satisfy the \emph{Equal Similarity} and \emph{High dimensionality} properties. Inspired by this, we propose \textbf{Dense Hash Encoding}, which seeks to combine the advantages of the above encodings and satisfy all the properties.

Without loss of generality, we assume feature values are integers as we can map string values to integers with string hashing\footnote{There is basically no collision due to the large output space ($2^{64} \approx10^{19}$ values for 64-bit integers). An example is \emph{CityHash64}: \url{https://github.com/google/cityhash}}. The proposed encoding function $E:\mathbb{N}\to \mathbb{R}^k$ uses $k$ universal hash functions to map a feature value to a $k$-dimensional dense and real-valued encodings. Specifically, we have $E'(s)=[H^{(1)}(s),H^{(2)}(s),\dots,H^{(k)}(s)]$ where $H^{(i)}:\mathbb{N}\to\{1,2,\dots,m\}$. Note that $m$ in this case is not related to the embedding table, and we just need to set it to a relatively large number ($10^6$ in our experiments). A nice property of universal hashing~\cite{DBLP:conf/stoc/CarterW77} is that the hashed values are evenly distributed (on average) over \{1,2,\dots,m\}. 

However, the integer-based $E'(s)$ encoding is not suitable to be used as the input to neural networks, as the input is typically real-valued and normalized for numeric stability. So we obtain real-valued encodings via appropriate transformations: $E(s)=\text{transform}(E'(s))$. We consider to approximate one of the following commonly used distributions:

\begin{itemize}[leftmargin=4mm]
\item\textbf{Uniform Distribution.} We simply normalize the encoding $E'$ to the range of $[-1,1]$. As the hashing values are evenly distributed (on average) among $\{1,2,\dots,m\}$, this approximates the uniform distribution $U(-1,1)$ reasonably well with a large $m$.
\item\textbf{Gaussian Distribution.} We first use the above transformation to obtain uniform distributed samples, and then apply the Box–Muller transform~\cite{box1958note} which converts the uniformly distributed samples~(i.e., $U(-1,1)$) to the standard normal distribution $\mathcal{N}(0, 1)$.  Please refer to Appendix for the implementation.
\end{itemize}

The choice of the two distributions is partially inspired by Generative Adversarial Networks~(GANs)~\cite{GAN} that typically draw random noise from a uniform or Gaussian distribution, and then feed it into neural networks for image generation. Note that the transformation (and the hashing) is deterministic, meaning the encoding (for any feature value) doesn't change over time. Empirically we found the two distributions work similarly well, and thus we choose the uniform distribution by default for simplicity.

Unlike existing hashing methods limited to a few hash functions, we choose a relatively large $k$ for satisfying the \emph{high-dimensionality} property ($k$=1024 in our experiments, though it's significantly smaller than $n$). We empirically found our method significantly benefits from larger $k$ while existing hashing methods do not. Moreover, the proposed dense hash encodings also satisfy the other three properties. More analysis can be found in Appendix.

Note that the whole encoding process does not require any storage, as all the computation can be done on the fly. This is also a nice property of using multiple hashing, as we obtain a more distinguishable higher-dimensional encoding without storage overhead. Computation-wise, the time complexity is $O(k)$ and calculation of each hashing is independent and thus amenable for parallelization and hardwares like GPUs and TPUs. As an example, we use the universal hashing for integers~\cite{DBLP:conf/stoc/CarterW77} as the underlying hashing, and depict the encoding process in Algorithm~\ref{algo:encod} in Appendix. Other universal hashing (e.g. for strings) could also be adopted for handling different feature types.

\subsection{Deep Embedding Network}

In DHE, the decoding function $F:\mathbb{R}^k\to\mathbb{R}^d$ needs to transform a $k$-dim encoding vector to a $d$-dim embedding. Obviously, the real-valued encoding is not applicable for embedding lookup. However, the mapping process is very similar to a highly non-linear feature transformation, where the input feature is fixed and non-learnable. Therefore, we use powerful deep neural networks~(DNN) to model such a complex transformation, as DNNs are expressive universal function approximators~\cite{DBLP:conf/nips/LuPWH017}. Moreover, a recent study shows that deeper networks are more parameter-efficient than shallow networks~\cite{DBLP:conf/iclr/LiangS17}, and thus DHE may reduce the model size compared against one-hot full embedding (a 1-layer shallow network).

However, the transformation task is highly challenging, even with DNNs. Essentially, the DNN needs to memorize the information (previously stored in the huge embedding table) in its weights. We hope the hierarchical structures and non-linear activations enable DNNs to express the embedding function more efficiently than the one-hot encodings (i.e., 1-layer wide NN). This is motivated by recent research that shows that deep networks can approximate functions with much fewer parameters compared with wide and shallow networks~\cite{DBLP:conf/iclr/LiangS17}.

Specifically, we use a feedforward network as the decoding function for DHE. We transform the $k$-dim encoding via $h$ hidden layers with $d_\text{NN}$ nodes. Then, the outputs layer (with $d$ nodes) transforms the last hidden layer to the $d$-dim feature value embedding. In practice, $d_{\text{NN}}$ is determined by the budget of memory consumption. We can see that the number of parameters in the DNN is $O(k * d_{\text{NN}} + (h-1) * d_{\text{NN}}^2+d_{\text{NN}}*d)$, which is independent of $n$ or $m$. This is also the time complexity of DHE. A unique feature of DHE is that it does not use any embedding table lookup, while purely relies on hidden layers to memorize and compute embeddings on the fly.

However, we found that training the deep embedding network is quite challenging (in contrast, one-hot based shallow networks are much easier to train). We observed inferior training and testing performance, presumably due to trainability and expressiveness issues. 
The expressiveness issue is relatively unique to our task, as NNs are often considered to be highly expressive and easy to overfit. However, we found the embedding network is underfitting instead of overfitting in our case, as the embedding generation task requires highly non-linear transformations from hash encodings to embeddings. We suspect the default ReLU activation ($f(x)$=$\max(0, x)$) is not expressive enough, as ReLU networks are piece-wise linear functions~\cite{DBLP:conf/iclr/AroraBMM18}. We tried various activation functions\footnote{we also tried tanh and SIREN~\cite{siren}, and found them on par or inferior to ReLU.} and found the recently proposed Mish activation~\cite{mish} ($f(x)$=$x\cdot\tanh(\ln(1+e^x))$) consistently performs better than ReLU and others. We also found batch normalization (BN)~\cite{bn} can stabilize the training and achieve better performance. However, regularization methods like dropout are not beneficial, which again verifies the bottleneck of our embedding network is underfitting. 


\subsection{Side Features Enhanced Encodings for Generalization}

An interesting extension for DHE utilizes side features for learning better encodings. This helps to inject structure into our encodings, and enable better generalization among feature values, and to new values.
One significant challenge of embedding learning for categorical features is that we can only \emph{memorize} the information for each feature value while we cannot \emph{generalize} among feature values~(e.g. generalize from ID 7 to ID 8) or to new values. This is due to the fact that the underlying feature representation does not imply any inherent similarity between different IDs. 
A typical method for achieving generalization is using side features which provide inherit similarities for generalization (e.g. dense features or bag-of-words features). However, these features are usually used as additional features for the recommendation model, and not used for improving embedding learning of the categorical feature.

One-hot based full embeddings inherit the property of categorical features, and generate the embeddings independently (i.e., the embeddings for any two IDs are independent). Thus, one-hot based schemes can be viewed as \textbf{decentralized} architectures that are good at \emph{memorization} but fail to achieve \emph{generalization}. In contrast, the DHE scheme is a \textbf{centralized} solution: any weight change in the embedding network will affect the embeddings for all feature values. We believe the centralized structure provides a potential opportunity for generalization.

As the decoding function of DHE is a neural network, we have a great flexibility to modify the input, like incorporating side features. We propose side feature enhanced encodings for DHE, and hope this will improve the generalization among feature values, and to new values. 
One straightforward way to enhance the encoding is directly concatenating the generalizable features and the hash encodings. If the dimensionality of the feature vector is too high, we could use locality-sensitive hashing~\cite{lsh} to significantly reduce the cardinality while preserving the tag similarity. The enhanced encoding is then fed into the deep embedding network for embedding generation. We think that the hash encoding provides a unique identifier for \emph{memorization} while the other features enable the \emph{generalization} ability.


\subsection{Summary} 

To wrap up, DHE has two major components: 1. dense hash encoding and 2. deep embedding network. The encoding module is fully deterministic and non-learnable, which doesn't require any storage for parameters. The embedding network transforms the identifier vector to the desired embedding. A significant difference of DHE is the absence of the embedding table, as we store all embedding information in the weights of DNN. Hence, all the feature values share the whole embedding network for embedding generation, unlike the hashing trick that shares the same embedding vector for different feature values. This makes DHE free of the embedding collision issue. The bottleneck of DHE is the computation of the embedding network, though it could be accelerated by more powerful hardware and NN acceleration approaches like pruning~\cite{DBLP:conf/iclr/FrankleC19}. 

Unlike existing embedding methods that explicitly assign each ID the same embedding length, an interesting fact of DHE is that the embedding capacity for each feature value is implicit and could be variable. This could be a desired property for online learning settings (the vocab size constantly grows) and power-law distributions (the embedding network may spend more parameters to memorize popular IDs).





\section{Experiments}

We conduct extensive experiments to investigate the following research questions:
\begin{itemize}[leftmargin=4mm]
    \item \textbf{RQ1:} How does DHE compare against full embedding and hash embedding methods that are based on embedding tables?
    \item \textbf{RQ2:} What's the effect of various encoding schemes for DHE?
    \item \textbf{RQ3:} How does the number of hash functions $k$ affect DHE and other hashing methods?
    \item \textbf{RQ4:} What's the influence of different embedding network architectures, depths, normalization and activation functions?
    \item \textbf{RQ5:} What's the effect of the side feature enhanced encoding?
    \item \textbf{RQ6:} What's the efficiency and GPU acceleration effect of DHE?
\end{itemize}








\begin{figure*}
\centering
\includegraphics[width=\linewidth]{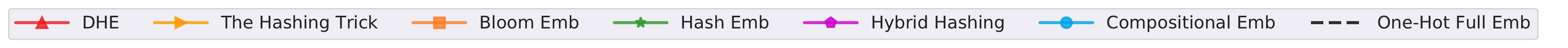}
\begin{subfigure}[b]{0.247\linewidth}
\includegraphics[width=\linewidth]{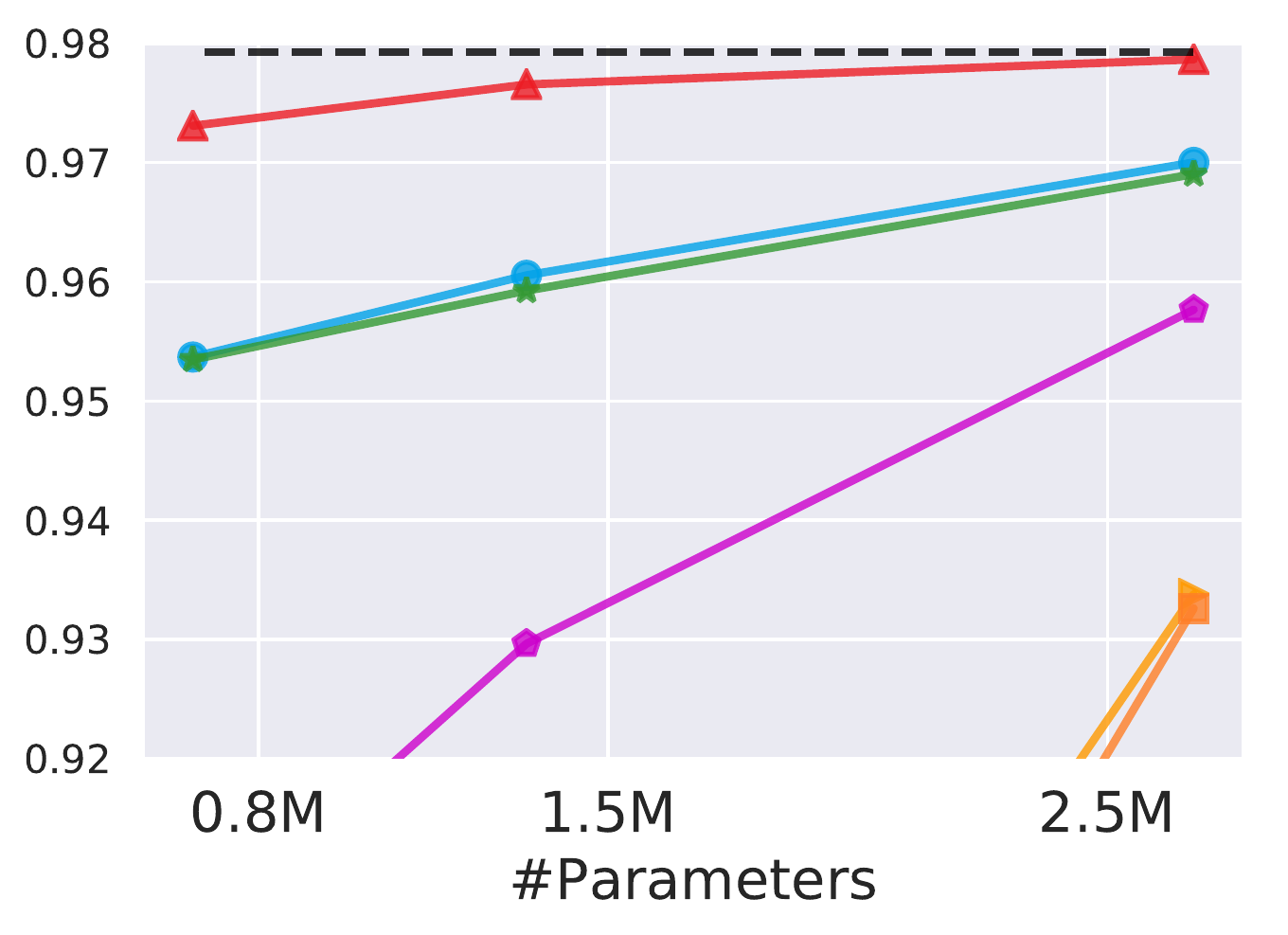}
\subcaption{ML-20M, GMF}
\end{subfigure}
\begin{subfigure}[b]{0.247\linewidth}
\includegraphics[width=\linewidth]{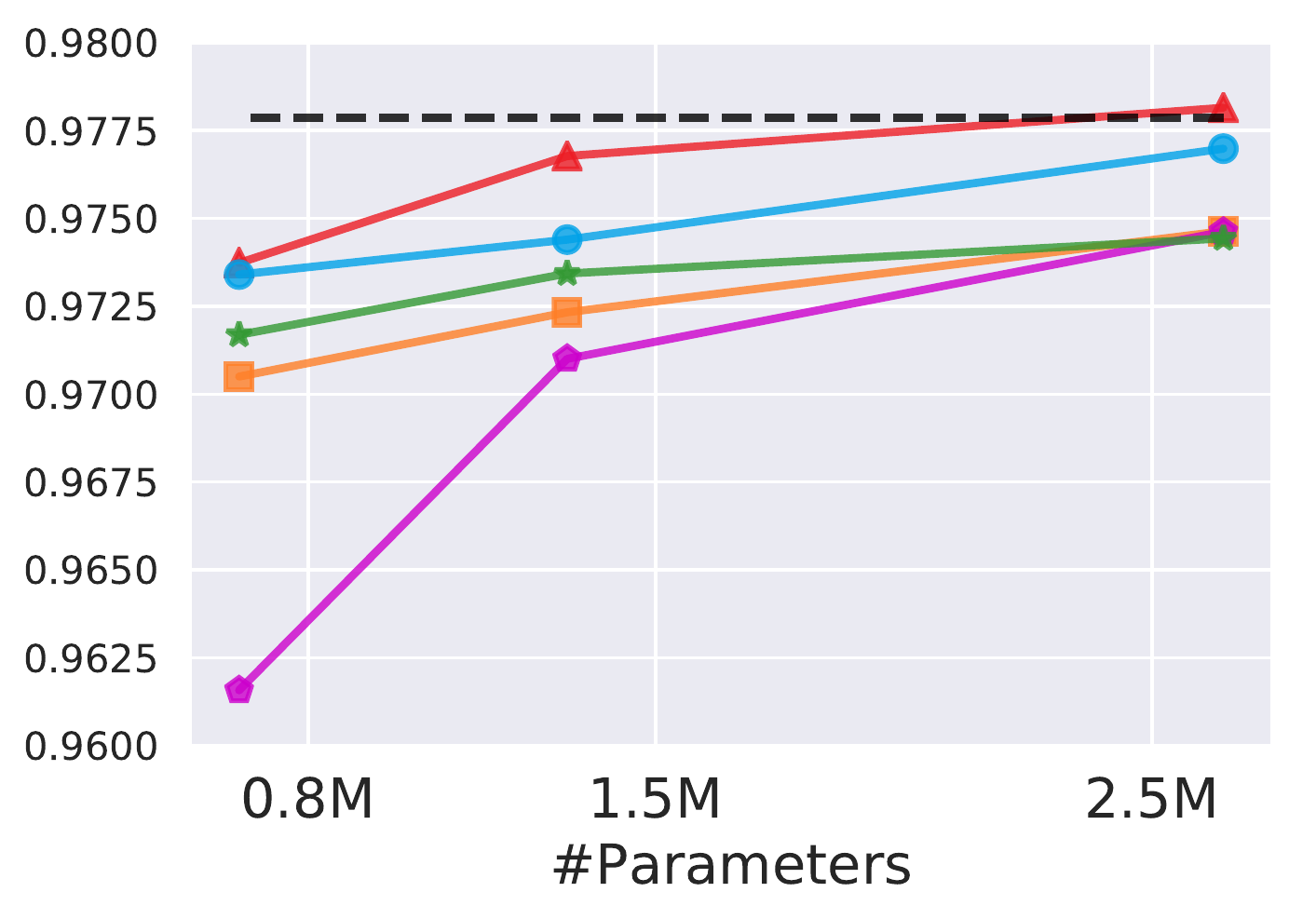}
\subcaption{ML-20M, MLP}
\end{subfigure}
\begin{subfigure}[b]{0.245\linewidth}
\includegraphics[width=\linewidth]{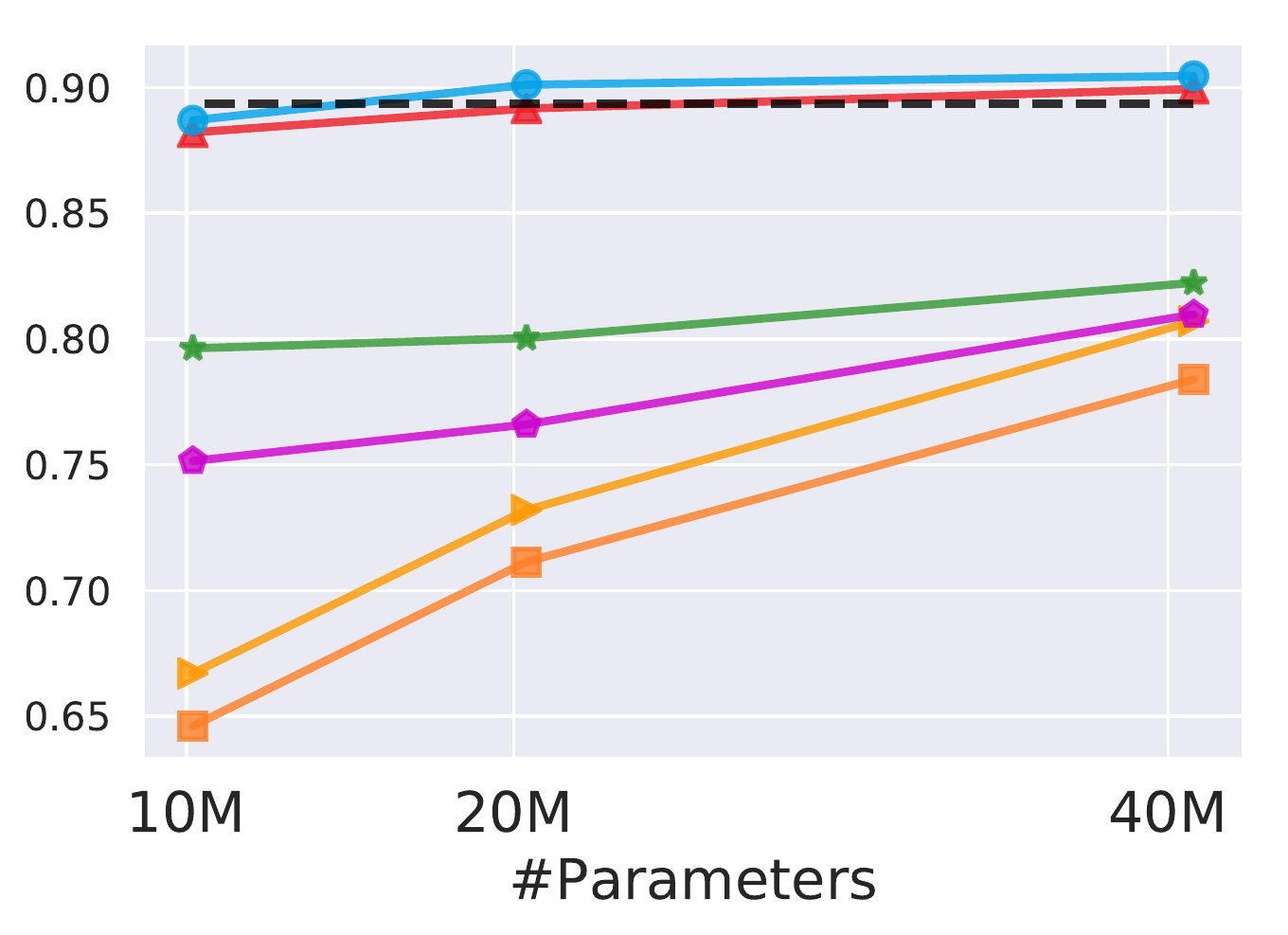}
\subcaption{Amazon, GMF}
\end{subfigure}
\begin{subfigure}[b]{0.245\linewidth}
\includegraphics[width=\linewidth]{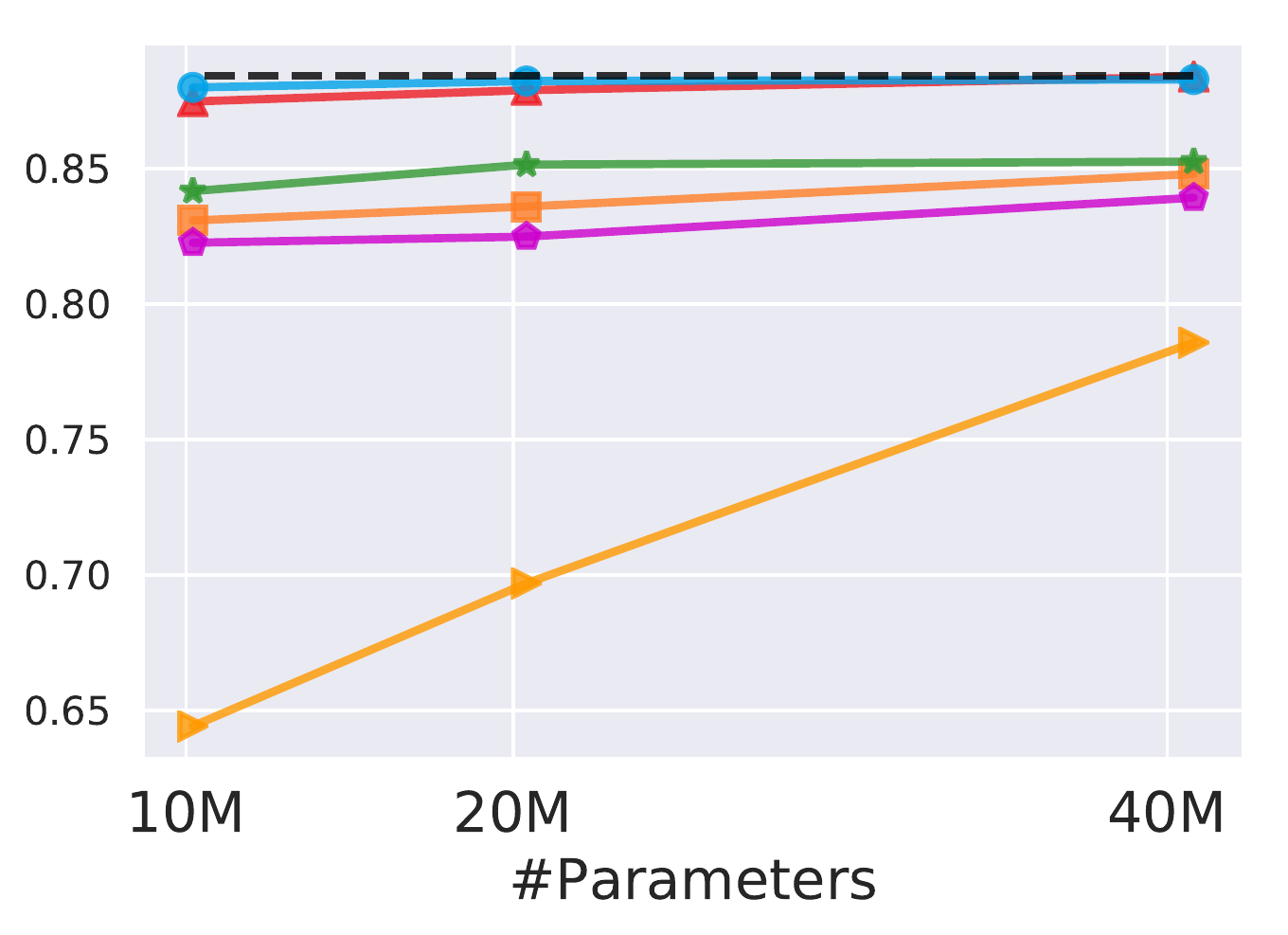}
\subcaption{Amazon, MLP}
\end{subfigure}
\vspace{-0.3cm}
\caption{AUC with different model sizes. The full embedding costs about 5M parameters for \emph{ML-20M}, and 80M for \emph{Amazon}.}\label{fig:ml}
\vspace{-0.3cm}
\end{figure*}

\subsection{Experimental Setup}

\subsubsection{Dataset}

We use two commonly used public benchmark datasets for evaluating recommendation performance:
\textbf{Movielens-20M}~\cite{DBLP:journals/tiis/HarperK16} and \textbf{Amazon Books}~\cite{DBLP:conf/emnlp/NiLM19}. The total vocab size (number of users and items) is 165K for \emph{Movielens-20M}, and 2.6M for \emph{Amazon Books}.  To reduce the variance, all the results are the average of the outcomes from 5 experiments. The results are on Movielens with 1/4 of the full model size, unless otherwise stated. More details are in Appendix.
 
\subsubsection{Evaluation Metric}
 We use AUC to evaluate recommendation performance.  AUC is a widely used metric in recommendation~\cite{DBLP:conf/uai/RendleFGS09, DBLP:conf/recsys/HeKM17} and CTR prediction~\cite{DBLP:conf/ijcai/GuoTYLH17}. The AUC measures the probability of ranking pairs of a positive item and negative items in the right order, and thus random guesses achieve an AUC of 0.5.

\subsubsection{Backbone Recommendation Models}

We adopt the \textbf{Generalized Matrix Factorization~(GMF)} and \textbf{Multi-layer Perceptron~(MLP)} from~\cite{DBLP:conf/www/HeLZNHC17} as the backbone recommendation models to evaluate the performance of different embedding approaches. We use the two methods to represent both shallow and deep recommendation models. GMF is a shallow model that calculates a weighted sum on  the element-wise product of user and item embeddings. With equal weights, GMF is reduced to the classic MF method. Conversely, MLP is a deep model that applies several fully-connected layers on the concatenation of user and item embeddings. Similar deep models have been adopted for recommendation and CTR prediction~\cite{DBLP:conf/ijcai/GuoTYLH17}. The MLP we used has three hidden layers (with [256, 128, 64] nodes), and an output layer to generate the $d$-dim embedding.




\subsection{Baselines}

The one-hot \textbf{Full Embedding} is a standard way to handle categorical features, which uses a dictionary to map each feature value to a unique one-hot vector. However, to adapt to online learning settings where new items constantly appear and stale items gradually vanish, or to reduce storage cost, hashing-based methods are often adopted. We use the follow hashing-based baselines:
\begin{itemize}[leftmargin=4mm]
\item\textbf{The Hashing Trick~\cite{hashing}} A classical approach for handling large-vocab categorical features, which uses a single hash function to map feature value into a smaller vocab. The method often suffers from collision problems.

\item\textbf{Bloom Embedding~\cite{bloom}} Inspired by bloom filter~\cite{DBLP:journals/cacm/Bloom70}, Bloom Embedding generates a binary encoding with multiple hash functions. Then a linear layer is applied to the encoding to recover the embedding for the given feature value.

\item\textbf{Hash Embedding (HashEmb)~\cite{hashemb}} HashEmb uses multiple (typically two) hash functions and lookups the corresponding embeddings. Then a weighted sum of the embeddings is adopted, where the weights are learned and dedicated for each feature value. 

\item\textbf{Hybrid Hashing~\cite{twitter}} A recently proposed method uses one-hot full embedding for frequent feature values, and uses double hashing for others.

\item\textbf{Compositional Embedding~\cite{fb}} A recently proposed method adopts two complementary hashing for avoiding hashing collision. We use the path-based version where the second hashing uses multiple MLPs with one hidden layer of 64 nodes.

\end{itemize}

We compare our \textbf{Deep Hashing Embedding~(DHE)} against the above baselines. DHE uses a large number of hash functions ($k$=1024 in the experiments) to generate a unique identifier for each feature value, followed by a deep embedding network to generate the final embedding. DHE also differs in that it doesn't use any one-hot encoding and embedding table lookup.


\subsection{Performance Comparison (RQ1)}

We plot the performance with 1/2, 1/4, and 1/8 of the full model size in Figure~\ref{fig:ml} for the two datasets. We interpret the results via the following comparisons:
\begin{itemize}[leftmargin=4mm]
\item{\emph{DHE vs. one-hot full emb:}} We observed that DHE effectively approximates Full Embedding's performance. In most cases, DHE achieves similar AUC with only 1/4 of the full model size. This verifies the effectiveness and efficiency (in model sizes) of DHE's hash encoding and deep embedding network, and shows that it's possible to remove one-hot encodings and embedding tables without AUC loss. 
\item{\emph{DHE vs. hashing methods:}} We can see that DHE significantly outperforms hashing-based baselines in most cases. This is attributed to its unique hash encoding, which is free of collision and easy for the embedding network to distinguish, and the expressive deep structures for embedding generation. The only exception is Compositional Embedding, which performs slightly better than DHE on the highly sparse \emph{Amazon} dataset. The Hash Trick~\cite{hashing} performs inferior to other methods, especially when the model size is small. This shows that the hash collisions severely hurt the performance.

\end{itemize}

\subsection{Comparison of Encoding Schemes (RQ2)}

We investigate the effect of various encodings (not based on one-hot) that are suitable for DHE. We evaluate DHE with encodings mentioned in Section~\ref{sec:encoding}: identity encoding (1-dim, normalized into [0,1]), binary encoding, random Fourier feature encoding~\cite{fourier}, and our proposed hashing encodings with the uniform or Gaussian distribution, and the results are shown in Table~\ref{exp:encod}. We can see that our proposed dense hash encoding with the uniform distribution is the best performer, while the Gaussian distribution variant is the runner-up. The binary encoding performs slightly inferior, and we think it is due to its wrong inductive bias (some IDs have more similar encodings) and the relatively low dimensionality~(i.e., $\left\lceil \log(n)\right\rceil$). The results also suggest that the random Fourier features~\cite{fourier} are not suitable for our case due to the difference between our problem and signal processing problems where the latter has a meaningful underlying temporal signal. This verifies the effectiveness of the dense hash encoding which satisfies the four properties we proposed.

\vspace{-0.1cm}
\begin{table}[htb]
\caption{AUC of DHE with different dense encoding schemes.}\label{exp:encod}\vspace{-0.3cm}
\centering\small
\setlength{\tabcolsep}{8pt}
\begin{tabularx}{.95\linewidth}{Xcc}
\toprule
\textbf{Encoding}                                    &\textbf{GMF} & \textbf{MLP} \\   \midrule
Identity Encoding                            & 94.99 &  95.01 \\ 
Binary Encoding                               & 97.38 & 97.36 \\
Random Fourier Encoding~\cite{fourier}          & 94.98 & 94.99 \\           
\textbf{Dense Hash Encoding-Gaussian} (ours)                             & 97.62 & 97.66  \\
\textbf{Dense Hash Encoding-Uniform} (ours)                           & \textbf{97.66} & \textbf{97.68}  \\\bottomrule
\end{tabularx}
\vspace{-0.2cm}
\end{table}

\subsection{Scalability Regarding the Number of Hash Functions (RQ3)}

Both our method DHE and multiple hashing based methods utilize multiple hash functions to reduce collision. However, as existing hashing methods are limited to a few hash functions (typically 2)~\cite{hashemb, twitter}, we investigate the scalability of DHE and the hashing baselines, in terms of the number of hash functions. Table~\ref{exp:nhash} shows the performance with different $k$, the number of hash functions. Note that the encoding length of DHE is $k$, the same as the number of hash functions, while the encoding length for one-hot hashing based methods is $m*k$.

With a small $k$ (e.g. $k\leq$8), the performance of DHE is inferior to the baselines, mainly because of the shorter encoding length of DHE~(i.e., $k$ versus $m*k$ for others). However, when $k\geq 32$, DHE is able to match or beat the performance of alternative methods. When $k$ further increases to more than 100, we can still observe performance gains of DHE, while the one-hot hashing baselines don't benefit from more hash functions. We suspect the reason for the poor utilization of multiple hashing is that each embedding will be shared $k$ times more than single hashing (if sharing embedding tables), and this leads to more collisions. If creating $k$ embedding tables (i.e., not sharing), given the same memory budget, the size for each table will be $k$ times smaller, which again causes the collision issue. However, DHE is free of the collision and embedding sharing problems, and thus can scale to a large $k$.

\vspace{-0.1cm}
\begin{table}[h]
\centering\small
\caption{The effect of the number of hash functions. The results are the AUC of the MLP recommendation model . ``-'' means the setting is infeasible for the memory budget.}\label{exp:nhash}\vspace{-0.3cm}
\setlength{\tabcolsep}{3pt}
\begin{tabular}{lccccccc}
\toprule
\textbf{\#hash functions ($k$)}			&	\textbf{2}	&	\textbf{4}	&	\textbf{8}	& \textbf{32} &	\textbf{128}	&	\textbf{1024} & \textbf{2048}\\ \midrule                               
Bloom Emb~\cite{bloom}  	&97.21&97.34&97.35&97.43&97.43&97.39&97.28\\
Hybrid Emb~\cite{twitter}   &97.20&97.31&\textbf{97.36}&97.42&97.42&97.41&97.30\\
Hash Emb~\cite{twitter}  					&	 \textbf{97.29} & \textbf{97.40}	&	- 	&	- 	&	- 	&	- 	&	- 	\\
DHE 	&			92.74&95.27&96.77&\textbf{97.44}&\textbf{97.58}&\textbf{97.67}&\textbf{97.65} \\                      
\bottomrule
\end{tabular}
\vspace{-0.1cm}
\label{tab:scale}
\end{table}

\subsection{Normalization and Activation (RQ4)}

Training the deep embedding network is much harder than training embedding lookup based shallow methods. We found there is a trainability issue as well as a unique expressiveness issue. We found that Batch Normalization~(BN)~\cite{bn} greatly stabilizes and accelerates the training, and improves the performance. For the expressiveness issue, we tried various activation functions for replacing ReLU, as ReLU networks are piece-wise linear functions~\cite{DBLP:conf/iclr/AroraBMM18} which may not be suitable for the complex transformation in our task. We found the recently proposed Mish~\cite{mish} activation is superior.

Table~\ref{tab:act} shows the results of with and without BN and Mish. We omit results of other activation functions, as we didn't observe performance improvement. We can see that both BN and Mish are critical for enabling deep networks for embedding generation, and improving DHE's performance. Note that for fair comparison, we only use BN and Mish for the embedding network in DHE, while use the same recommendation model (e.g. the MLP model) for all embedding methods.

\vspace{-0.1cm}
\begin{table}[htb]
\caption{The effect of activations functions and normalization.}\label{tab:act}
\centering\small\vspace{-0.3cm}
\setlength{\tabcolsep}{10pt}
\begin{tabularx}{.8\linewidth}{Xlll}
\toprule
\textbf{Activation functions}                                    &\textbf{GMF} & \textbf{MLP}  \\  \midrule
\emph{Without Batch Normalization}\\
ReLU             & 97.33 &97.47  \\
Mish~\cite{mish}             & 97.43&  97.50 \\ \hline
\emph{With Batch Normalization}\\
ReLU             & 97.54 &97.59  \\
Mish~\cite{mish} (default)       &\textbf{97.66} & \textbf{97.67} \\ \bottomrule
\end{tabularx}
\vspace{-0.2cm}
\end{table}

\subsection{The Effect of Depth (RQ4)}

\begin{figure}[htb]
\centering
\includegraphics[width=.6\linewidth]{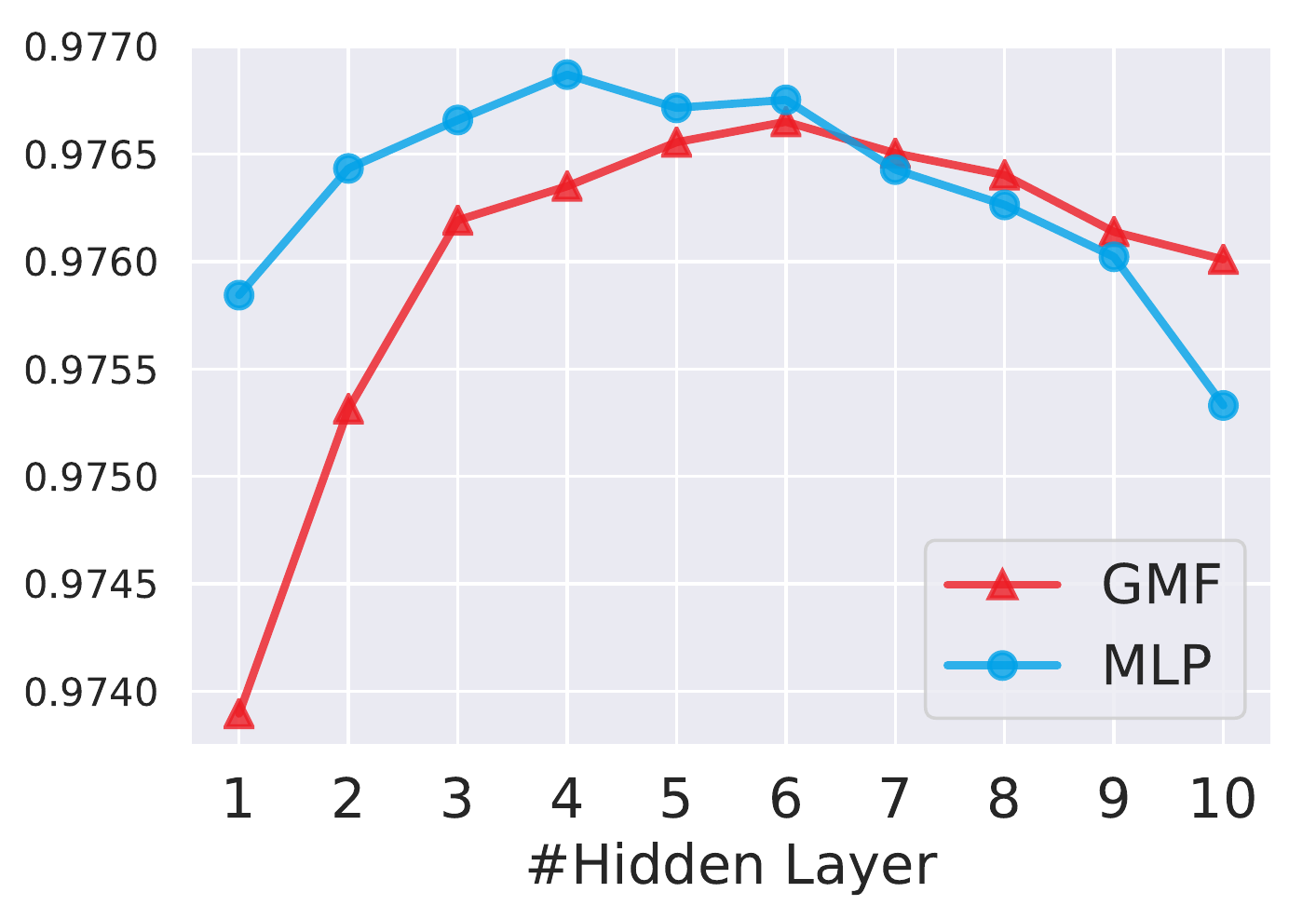}
\vspace{-0.3cm}
\caption{AUC with different depths on Movielens. All the data points are with the same \#params. The network depth is \#Hidden Layer plus one, where the last layer is for generating the embedding.}\label{fig:depth}
\vspace{-0.2cm}
\end{figure}

The embedding network in DHE takes a hash encoding vector and applies a deep neural network to generate the $d$-dim output embedding. Specifically, the embedding network consists of several hidden layers with $d_{\text{NN}}$ nodes, followed by an output layer with $d$ nodes. We investigate whether deeper embedding networks are more effective against wide \& shallow networks, via varying the number of hidden layers while keeping the same number of parameters. Figure~\ref{fig:depth} shows the results on Movielens. We observed that embedding networks with around five hidden layers are significantly better than wider and shallower networks. This is consistent with our motivation and theoretical results in~\cite{DBLP:conf/iclr/LiangS17}, that deep networks are more parameter-efficient than shallow networks. However, we didn't see further improvement with more hidden layers, presumably because each layer's width is too narrow or due to trainability issues on deep networks. 

\subsection{Neural Architectures (RQ4)}

The default neural architecture for DHE is equal-width MLP, where each hidden layer has $d_\text{NN}$ nodes. We also explore various architectures including \emph{Pyramid MLP} (the width of a hidden layer is twice that of the previous layer), \emph{Inverted Pyramid MLP} (opposite to Pyramid MLP), \emph{DenseNet~\cite{densenet}-like MLP} (concatenate all previous layers' output as the input at each layer), and equal-width MLP with residual connections~\cite{resnet}. We adjust the width to make sure all the variants have the same number of parameters. The performance results are shown in Table~\ref{tab:neural}. We can see that the simple equal-width MLP performs the best, and adding residual connections also slightly hurts the performance. We suspect that the low-level representations are not useful in our case, so that the attempts (as in computer vision) utilizing low-level features (like DenseNet~\cite{densenet} or ResNet~\cite{resnet}) didn't achieve better performance. The (inverted) Pyramid MLPs also perform worse than the equal-width MLP, perhaps more tuning on the width multiplier (we used 2 and 0.5) is needed. The results also show it's challenging to design architectures for the embedding generalization tasks, as we didn't find useful prior to guide our designs.

\vspace{-0.1cm}
\begin{table}[htb]
\caption{AUC with different neural architectures for the embedding network in DHE. All variants use 5 hidden layers with BN and Mish, and have the same number of parameter.}\label{tab:neural}\vspace{-0.3cm}
\centering\small
\setlength{\tabcolsep}{10pt}
\begin{tabularx}{.9\linewidth}{Xlllll}
\toprule
\textbf{Emb Network}                                    &\textbf{GMF} & \textbf{MLP}  \\  \midrule
Pyramid MLP             & 97.20 & 97.49  \\
Inverted Pyramid MLP             & 97.50 & 97.58  \\
DenseNet-like MLP             & 97.53 & 97.50 \\
Residual Equal-Width MLP             & 97.61 & 97.60  \\
Equal-Width MLP (default)            & \textbf{97.66} & \textbf{97.68}  \\ \bottomrule
\end{tabularx}
\vspace{-0.2cm}
\end{table}

\subsection{Side Feature Enhanced Encodings (RQ5)}

In previous experiments, we don't use side features in DHE for fair comparison. To investigate the effect of the side feature enhanced encoding, we use the 20 movie \emph{Genres}~(e.g. `Comedy', `Romance', etc.) in the Movielens dataset, as the side feature. Each movie has zero, one, or multiple genres, and we represent the feature with a 20-dim binary vector. The side features can be used in the encoding function of DHE, and/or directly plugged into the MLP recommendation model (i.e., the MLP takes user, item, and genres vectors as the input).

The results are shown in Table~\ref{tab:side}. We can see that using side features only in the encoding and only in the MLP have similar performance. This shows DHE's item embeddings effectively capture the \emph{Genres} information, and verifies the generalization ability of item embeddings generated by DHE with enhanced encodings. However, we didn't see further improvement of using the feature in both encoding and MLP. For other embedding methods, adding the feature to the MLP is helpful. However, unlike DHE, they fully rely on IDs and are unable to generate generalizable item embeddings. 

\vspace{-0.1cm}
\begin{table}[htb]
\caption{The effect of side feature enhanced encoding.}\label{exp:feature}\vspace{-0.3cm}\label{tab:side}
\centering\small
\begin{tabularx}{\linewidth}{Xcc}
\toprule
\textbf{Item Embedding} &\textbf{MLP (ID only)}&\textbf{MLP (with genres)}\\
 \midrule
One-hot Full Emb &97.64 &97.67 \\
\midrule
\emph{DHE Encoding}\\
hash encoding (ID) only     & 97.67   &97.72   \\
Genres only & 79.17 & 79.16\\
hash encoding (ID) +Genres   & \textbf{97.71} &\textbf{97.73} \\
\midrule
Hash Emb &97.34 &97.42\\ 
Hybrid Hashing&97.22&97.31\\
\bottomrule
\end{tabularx}
\vspace{-0.2cm}
\end{table}


\subsection{Efficiency (RQ6)}

One potential drawback of DHE is computation efficiency, as the neural network module in DHE requires a lot of computing resources. However, this is a common problem in all deep learning models, and we hope the efficiency issue could be alleviated by powerful computation hardware (like GPUs and TPUs, optimized for neural networks) that are improving very fast recently. We show the efficiency results in Table~\ref{exp:time}. With GPUs, DHE is about 9x slower than full embedding, and 4.5x slower than hash embeddings. However, we can see that DHE significantly benefits from GPU acceleration, while full embeddings don't. This is because the embedding lookup process in full embeddings is hard to accelerate by GPUs. The result conveys a promising message that more powerful computation hardware in the future could further accelerate DHE, and gradually close the efficiency gap. Moreover, DHE could also potentially benefit from NN acceleration methods, like pruning~\cite{DBLP:conf/iclr/FrankleC19}.

\vspace{-0.1cm}
\begin{table}[h]
\centering\small
\caption{Time (in seconds) of embedding generation for 1M queries with a batch size of 100.}\label{exp:time}\vspace{-0.3cm}
\setlength{\tabcolsep}{10pt}
\begin{tabular}{lrrr}
\toprule
			&	\textbf{CPU}	&	\textbf{GPU}	&	\textbf{GPU Acceleration}\\ \midrule  
Full Emb & 3.4& 3.4& -1\%\\
Hash Emb~\cite{hashemb} & 8.4& 6.1& -26\%\\
DHE      & 76.1&27.2&-64\%\\
\bottomrule
\end{tabular}
\vspace{-0.2cm}
\end{table}

\section{Related Work}

Embedding learning has been widely adopted, and representative examples include word2vec~\cite{word2vec} and Matrix Factorization~(MF)~\cite{DBLP:conf/uai/RendleFGS09}. Other than `shallow models,' embeddings are also the key component of deep models, like word embeddings for BERT~\cite{bert}. There are various work on improving the performance or efficiency of embedding learning, via dimensionality search~\cite{DBLP:conf/kdd/JoglekarLCXWAKL20}, factorization~\cite{DBLP:conf/iclr/LanCGGSS20}, pruning~\cite{pep}, etc. However, these methods are orthogonal to DHE as they are built on top of the standard one-hot encoding and embedding tables.

The hashing trick~\cite{hashing} is a classic method enabling handling large-vocab features and out-of-vocab feature values with one-hot encodings. As only a single hash function is adopted, the collision issue becomes severe when the number of hashing buckets is small. To alleviate this, various improved methods~\cite{bloom, twitter, hashemb} are proposed based on the idea of using multiple hash functions to generate multiple one-hot encodings. Our method also adopts hash functions for generating the encoding. However, our method doesn't rely on one-hot encodings. Also, our approach is able to scale to use a large number of hash functions, while existing methods are limited to use a few (typically two) hash functions.

There is an orthogonal line of work using similarity-preserving hashing for embedding learning. For example, HashRec~\cite{DBLP:conf/cikm/KangM19} learns preference-preserving binary representation for efficient retrieval, where a low hamming distance between the embeddings of a user and an item indicates the user may prefer the item. Some other methods utilize locality-sensitive hashing~\cite{lsh} to reduce feature dimensions while maintaining their similarities in the original feature spaces~\cite{DBLP:conf/icml/Ravi19, DBLP:conf/emnlp/KrishnamoorthiR19}. The main difference is that the hashing we used are designed for reducing collision, while the hashing used in these methods seeks to preserve 
some kind of similarity.


\section{Conclusions and Future Work}

In this work, we revisited the widely adopted one-hot based embedding methods, and proposed an alternative embedding framework (DHE), based on dense hash encodings and deep neural networks. DHE does not lookup embeddings, and instead computes embeddings on the fly through its hashing functions and an embedding network. This avoids creating and maintaining huge embedding tables for training and serving.
As a DNN-based embedding framework, DHE could benefit significantly from future deep learning advancement in modeling and hardware, which will further improve DHE's performance and efficiency.

In the future, we plan to investigate several directions for extending and improving DHE: (i) handling multivalent features like bag-of-words; (ii) jointly modeling multiple features with DHE; (iii) hybrid approaches using both embedding tables and neural networks, for balancing efficiency and performance.


\bibliographystyle{ACM-Reference-Format}
\bibliography{DHE}


\begin{thebibliography}{38}


\ifx \showCODEN    \undefined \def \showCODEN     #1{\unskip}     \fi
\ifx \showDOI      \undefined \def \showDOI       #1{#1}\fi
\ifx \showISBNx    \undefined \def \showISBNx     #1{\unskip}     \fi
\ifx \showISBNxiii \undefined \def \showISBNxiii  #1{\unskip}     \fi
\ifx \showISSN     \undefined \def \showISSN      #1{\unskip}     \fi
\ifx \showLCCN     \undefined \def \showLCCN      #1{\unskip}     \fi
\ifx \shownote     \undefined \def \shownote      #1{#1}          \fi
\ifx \showarticletitle \undefined \def \showarticletitle #1{#1}   \fi
\ifx \showURL      \undefined \def \showURL       {\relax}        \fi
\providecommand\bibfield[2]{#2}
\providecommand\bibinfo[2]{#2}
\providecommand\natexlab[1]{#1}
\providecommand\showeprint[2][]{arXiv:#2}

\bibitem[\protect\citeauthoryear{Arora, Basu, Mianjy, and Mukherjee}{Arora
  et~al\mbox{.}}{2018}]%
        {DBLP:conf/iclr/AroraBMM18}
\bibfield{author}{\bibinfo{person}{Raman Arora}, \bibinfo{person}{Amitabh
  Basu}, \bibinfo{person}{Poorya Mianjy}, {and} \bibinfo{person}{Anirbit
  Mukherjee}.} \bibinfo{year}{2018}\natexlab{}.
\newblock \showarticletitle{Understanding Deep Neural Networks with Rectified
  Linear Units}. In \bibinfo{booktitle}{\emph{ICLR}}.
\newblock


\bibitem[\protect\citeauthoryear{Bloom}{Bloom}{1970}]%
        {DBLP:journals/cacm/Bloom70}
\bibfield{author}{\bibinfo{person}{Burton~H. Bloom}.}
  \bibinfo{year}{1970}\natexlab{}.
\newblock \showarticletitle{Space/Time Trade-offs in Hash Coding with Allowable
  Errors}.
\newblock \bibinfo{journal}{\emph{Commun. {ACM}}} \bibinfo{volume}{13},
  \bibinfo{number}{7} (\bibinfo{year}{1970}), \bibinfo{pages}{422--426}.
\newblock


\bibitem[\protect\citeauthoryear{Box}{Box}{1958}]%
        {box1958note}
\bibfield{author}{\bibinfo{person}{George~EP Box}.}
  \bibinfo{year}{1958}\natexlab{}.
\newblock \showarticletitle{A note on the generation of random normal
  deviates}.
\newblock \bibinfo{journal}{\emph{Ann. Math. Stat.}}  \bibinfo{volume}{29}
  (\bibinfo{year}{1958}), \bibinfo{pages}{610--611}.
\newblock


\bibitem[\protect\citeauthoryear{Carter and Wegman}{Carter and Wegman}{1977}]%
        {DBLP:conf/stoc/CarterW77}
\bibfield{author}{\bibinfo{person}{Larry Carter} {and} \bibinfo{person}{Mark~N.
  Wegman}.} \bibinfo{year}{1977}\natexlab{}.
\newblock \showarticletitle{Universal Classes of Hash Functions (Extended
  Abstract)}. In \bibinfo{booktitle}{\emph{STOC}}.
\newblock


\bibitem[\protect\citeauthoryear{Covington, Adams, and Sargin}{Covington
  et~al\mbox{.}}{2016}]%
        {DBLP:conf/recsys/CovingtonAS16}
\bibfield{author}{\bibinfo{person}{Paul Covington}, \bibinfo{person}{Jay
  Adams}, {and} \bibinfo{person}{Emre Sargin}.}
  \bibinfo{year}{2016}\natexlab{}.
\newblock \showarticletitle{Deep Neural Networks for YouTube Recommendations}.
  In \bibinfo{booktitle}{\emph{RecSys}}. \bibinfo{publisher}{{ACM}},
  \bibinfo{pages}{191--198}.
\newblock


\bibitem[\protect\citeauthoryear{Datar, Immorlica, Indyk, and Mirrokni}{Datar
  et~al\mbox{.}}{2004}]%
        {lsh}
\bibfield{author}{\bibinfo{person}{Mayur Datar}, \bibinfo{person}{Nicole
  Immorlica}, \bibinfo{person}{Piotr Indyk}, {and} \bibinfo{person}{Vahab~S.
  Mirrokni}.} \bibinfo{year}{2004}\natexlab{}.
\newblock \showarticletitle{Locality-sensitive hashing scheme based on p-stable
  distributions}. In \bibinfo{booktitle}{\emph{SoCG}}.
  \bibinfo{publisher}{{ACM}}, \bibinfo{pages}{253--262}.
\newblock


\bibitem[\protect\citeauthoryear{Devlin, Chang, Lee, and Toutanova}{Devlin
  et~al\mbox{.}}{2019}]%
        {bert}
\bibfield{author}{\bibinfo{person}{Jacob Devlin}, \bibinfo{person}{Ming{-}Wei
  Chang}, \bibinfo{person}{Kenton Lee}, {and} \bibinfo{person}{Kristina
  Toutanova}.} \bibinfo{year}{2019}\natexlab{}.
\newblock \showarticletitle{{BERT:} Pre-training of Deep Bidirectional
  Transformers for Language Understanding}. In
  \bibinfo{booktitle}{\emph{NAACL-HLT}}. \bibinfo{publisher}{Association for
  Computational Linguistics}.
\newblock


\bibitem[\protect\citeauthoryear{Frankle and Carbin}{Frankle and
  Carbin}{2019}]%
        {DBLP:conf/iclr/FrankleC19}
\bibfield{author}{\bibinfo{person}{Jonathan Frankle} {and}
  \bibinfo{person}{Michael Carbin}.} \bibinfo{year}{2019}\natexlab{}.
\newblock \showarticletitle{The Lottery Ticket Hypothesis: Finding Sparse,
  Trainable Neural Networks}. In \bibinfo{booktitle}{\emph{ICLR}}.
  \bibinfo{publisher}{OpenReview.net}.
\newblock


\bibitem[\protect\citeauthoryear{Goodfellow, Pouget{-}Abadie, Mirza, Xu,
  Warde{-}Farley, Ozair, Courville, and Bengio}{Goodfellow
  et~al\mbox{.}}{2014}]%
        {GAN}
\bibfield{author}{\bibinfo{person}{Ian~J. Goodfellow}, \bibinfo{person}{Jean
  Pouget{-}Abadie}, \bibinfo{person}{Mehdi Mirza}, \bibinfo{person}{Bing Xu},
  \bibinfo{person}{David Warde{-}Farley}, \bibinfo{person}{Sherjil Ozair},
  \bibinfo{person}{Aaron~C. Courville}, {and} \bibinfo{person}{Yoshua Bengio}.}
  \bibinfo{year}{2014}\natexlab{}.
\newblock \showarticletitle{Generative Adversarial Nets}. In
  \bibinfo{booktitle}{\emph{NIPS}}.
\newblock


\bibitem[\protect\citeauthoryear{Guo, Tang, Ye, Li, and He}{Guo
  et~al\mbox{.}}{2017}]%
        {DBLP:conf/ijcai/GuoTYLH17}
\bibfield{author}{\bibinfo{person}{Huifeng Guo}, \bibinfo{person}{Ruiming
  Tang}, \bibinfo{person}{Yunming Ye}, \bibinfo{person}{Zhenguo Li}, {and}
  \bibinfo{person}{Xiuqiang He}.} \bibinfo{year}{2017}\natexlab{}.
\newblock \showarticletitle{DeepFM: {A} Factorization-Machine based Neural
  Network for {CTR} Prediction}. In \bibinfo{booktitle}{\emph{IJCAI}}.
  \bibinfo{publisher}{ijcai.org}.
\newblock


\bibitem[\protect\citeauthoryear{Harper and Konstan}{Harper and
  Konstan}{2016}]%
        {DBLP:journals/tiis/HarperK16}
\bibfield{author}{\bibinfo{person}{F.~Maxwell Harper} {and}
  \bibinfo{person}{Joseph~A. Konstan}.} \bibinfo{year}{2016}\natexlab{}.
\newblock \showarticletitle{The MovieLens Datasets: History and Context}.
\newblock \bibinfo{journal}{\emph{{ACM} Trans. Interact. Intell. Syst.}}
  \bibinfo{volume}{5}, \bibinfo{number}{4} (\bibinfo{year}{2016}),
  \bibinfo{pages}{19:1--19:19}.
\newblock


\bibitem[\protect\citeauthoryear{He, Zhang, Ren, and Sun}{He
  et~al\mbox{.}}{2016}]%
        {resnet}
\bibfield{author}{\bibinfo{person}{Kaiming He}, \bibinfo{person}{Xiangyu
  Zhang}, \bibinfo{person}{Shaoqing Ren}, {and} \bibinfo{person}{Jian Sun}.}
  \bibinfo{year}{2016}\natexlab{}.
\newblock \showarticletitle{Deep Residual Learning for Image Recognition}. In
  \bibinfo{booktitle}{\emph{CVPR}}. \bibinfo{publisher}{{IEEE} Computer
  Society}, \bibinfo{pages}{770--778}.
\newblock


\bibitem[\protect\citeauthoryear{He, Kang, and McAuley}{He
  et~al\mbox{.}}{2017a}]%
        {DBLP:conf/recsys/HeKM17}
\bibfield{author}{\bibinfo{person}{Ruining He}, \bibinfo{person}{Wang{-}Cheng
  Kang}, {and} \bibinfo{person}{Julian~J. McAuley}.}
  \bibinfo{year}{2017}\natexlab{a}.
\newblock \showarticletitle{Translation-based Recommendation}. In
  \bibinfo{booktitle}{\emph{RecSys}}. \bibinfo{publisher}{{ACM}}.
\newblock


\bibitem[\protect\citeauthoryear{He, Liao, Zhang, Nie, Hu, and Chua}{He
  et~al\mbox{.}}{2017b}]%
        {DBLP:conf/www/HeLZNHC17}
\bibfield{author}{\bibinfo{person}{Xiangnan He}, \bibinfo{person}{Lizi Liao},
  \bibinfo{person}{Hanwang Zhang}, \bibinfo{person}{Liqiang Nie},
  \bibinfo{person}{Xia Hu}, {and} \bibinfo{person}{Tat{-}Seng Chua}.}
  \bibinfo{year}{2017}\natexlab{b}.
\newblock \showarticletitle{Neural Collaborative Filtering}. In
  \bibinfo{booktitle}{\emph{WWW}}. \bibinfo{publisher}{{ACM}}.
\newblock


\bibitem[\protect\citeauthoryear{Huang, Liu, and Weinberger}{Huang
  et~al\mbox{.}}{2016}]%
        {densenet}
\bibfield{author}{\bibinfo{person}{Gao Huang}, \bibinfo{person}{Zhuang Liu},
  {and} \bibinfo{person}{Kilian~Q. Weinberger}.}
  \bibinfo{year}{2016}\natexlab{}.
\newblock \showarticletitle{Densely Connected Convolutional Networks}.
\newblock \bibinfo{journal}{\emph{CoRR}}  \bibinfo{volume}{abs/1608.06993}
  (\bibinfo{year}{2016}).
\newblock
\showeprint[arxiv]{1608.06993}


\bibitem[\protect\citeauthoryear{Ioffe and Szegedy}{Ioffe and Szegedy}{2015}]%
        {bn}
\bibfield{author}{\bibinfo{person}{Sergey Ioffe} {and}
  \bibinfo{person}{Christian Szegedy}.} \bibinfo{year}{2015}\natexlab{}.
\newblock \showarticletitle{Batch Normalization: Accelerating Deep Network
  Training by Reducing Internal Covariate Shift}. In
  \bibinfo{booktitle}{\emph{ICML}} \emph{(\bibinfo{series}{{JMLR} Workshop and
  Conference Proceedings}, Vol.~\bibinfo{volume}{37})}.
  \bibinfo{publisher}{JMLR.org}.
\newblock


\bibitem[\protect\citeauthoryear{Joglekar, Li, Chen, Xu, Wang, Adams, Khaitan,
  Liu, and Le}{Joglekar et~al\mbox{.}}{2020}]%
        {DBLP:conf/kdd/JoglekarLCXWAKL20}
\bibfield{author}{\bibinfo{person}{Manas~R. Joglekar}, \bibinfo{person}{Cong
  Li}, \bibinfo{person}{Mei Chen}, \bibinfo{person}{Taibai Xu},
  \bibinfo{person}{Xiaoming Wang}, \bibinfo{person}{Jay~K. Adams},
  \bibinfo{person}{Pranav Khaitan}, \bibinfo{person}{Jiahui Liu}, {and}
  \bibinfo{person}{Quoc~V. Le}.} \bibinfo{year}{2020}\natexlab{}.
\newblock \showarticletitle{Neural Input Search for Large Scale Recommendation
  Models}. In \bibinfo{booktitle}{\emph{SIGKDD}}. \bibinfo{publisher}{{ACM}}.
\newblock


\bibitem[\protect\citeauthoryear{Kang and McAuley}{Kang and McAuley}{2019}]%
        {DBLP:conf/cikm/KangM19}
\bibfield{author}{\bibinfo{person}{Wang{-}Cheng Kang} {and}
  \bibinfo{person}{Julian~John McAuley}.} \bibinfo{year}{2019}\natexlab{}.
\newblock \showarticletitle{Candidate Generation with Binary Codes for
  Large-Scale Top-N Recommendation}. In \bibinfo{booktitle}{\emph{CIKM}}.
  \bibinfo{publisher}{{ACM}}.
\newblock


\bibitem[\protect\citeauthoryear{Krishnamoorthi, Ravi, and
  Kozareva}{Krishnamoorthi et~al\mbox{.}}{2019}]%
        {DBLP:conf/emnlp/KrishnamoorthiR19}
\bibfield{author}{\bibinfo{person}{Karthik Krishnamoorthi},
  \bibinfo{person}{Sujith Ravi}, {and} \bibinfo{person}{Zornitsa Kozareva}.}
  \bibinfo{year}{2019}\natexlab{}.
\newblock \showarticletitle{{PRADO:} Projection Attention Networks for Document
  Classification On-Device}. In \bibinfo{booktitle}{\emph{EMNLP-IJCNLP}}.
  \bibinfo{publisher}{Association for Computational Linguistics},
  \bibinfo{pages}{5011--5020}.
\newblock


\bibitem[\protect\citeauthoryear{Lan, Chen, Goodman, Gimpel, Sharma, and
  Soricut}{Lan et~al\mbox{.}}{2020}]%
        {DBLP:conf/iclr/LanCGGSS20}
\bibfield{author}{\bibinfo{person}{Zhenzhong Lan}, \bibinfo{person}{Mingda
  Chen}, \bibinfo{person}{Sebastian Goodman}, \bibinfo{person}{Kevin Gimpel},
  \bibinfo{person}{Piyush Sharma}, {and} \bibinfo{person}{Radu Soricut}.}
  \bibinfo{year}{2020}\natexlab{}.
\newblock \showarticletitle{{ALBERT:} {A} Lite {BERT} for Self-supervised
  Learning of Language Representations}. In \bibinfo{booktitle}{\emph{ICLR}}.
  \bibinfo{publisher}{OpenReview.net}.
\newblock


\bibitem[\protect\citeauthoryear{Liang and Srikant}{Liang and Srikant}{2017}]%
        {DBLP:conf/iclr/LiangS17}
\bibfield{author}{\bibinfo{person}{Shiyu Liang} {and} \bibinfo{person}{R.
  Srikant}.} \bibinfo{year}{2017}\natexlab{}.
\newblock \showarticletitle{Why Deep Neural Networks for Function
  Approximation?}. In \bibinfo{booktitle}{\emph{ICLR}}.
  \bibinfo{publisher}{OpenReview.net}.
\newblock


\bibitem[\protect\citeauthoryear{Liu, Gao, Chen, Jin, and Li}{Liu
  et~al\mbox{.}}{2020}]%
        {pep}
\bibfield{author}{\bibinfo{person}{Siyi Liu}, \bibinfo{person}{Chen Gao},
  \bibinfo{person}{Yihong Chen}, \bibinfo{person}{Depeng Jin}, {and}
  \bibinfo{person}{Yong Li}.} \bibinfo{year}{2020}\natexlab{}.
\newblock \showarticletitle{Learnable Embedding Sizes for Recommender Systems}.
  In \bibinfo{booktitle}{\emph{ICLR}}.
\newblock


\bibitem[\protect\citeauthoryear{Lu, Pu, Wang, Hu, and Wang}{Lu
  et~al\mbox{.}}{2017}]%
        {DBLP:conf/nips/LuPWH017}
\bibfield{author}{\bibinfo{person}{Zhou Lu}, \bibinfo{person}{Hongming Pu},
  \bibinfo{person}{Feicheng Wang}, \bibinfo{person}{Zhiqiang Hu}, {and}
  \bibinfo{person}{Liwei Wang}.} \bibinfo{year}{2017}\natexlab{}.
\newblock \showarticletitle{The Expressive Power of Neural Networks: {A} View
  from the Width}. In \bibinfo{booktitle}{\emph{NIPS}}.
\newblock


\bibitem[\protect\citeauthoryear{Mikolov, Sutskever, Chen, Corrado, and
  Dean}{Mikolov et~al\mbox{.}}{2013}]%
        {word2vec}
\bibfield{author}{\bibinfo{person}{Tomas Mikolov}, \bibinfo{person}{Ilya
  Sutskever}, \bibinfo{person}{Kai Chen}, \bibinfo{person}{Gregory~S. Corrado},
  {and} \bibinfo{person}{Jeffrey Dean}.} \bibinfo{year}{2013}\natexlab{}.
\newblock \showarticletitle{Distributed Representations of Words and Phrases
  and their Compositionality}. In \bibinfo{booktitle}{\emph{NIPS}}.
\newblock


\bibitem[\protect\citeauthoryear{Misra}{Misra}{2010}]%
        {mish}
\bibfield{author}{\bibinfo{person}{Diganta Misra}.}
  \bibinfo{year}{2010}\natexlab{}.
\newblock \showarticletitle{Mish: {A} Self Regularized Non-Monotonic Neural
  Activation Function}. In \bibinfo{booktitle}{\emph{BMVC}}.
\newblock


\bibitem[\protect\citeauthoryear{Ni, Li, and McAuley}{Ni et~al\mbox{.}}{2019}]%
        {DBLP:conf/emnlp/NiLM19}
\bibfield{author}{\bibinfo{person}{Jianmo Ni}, \bibinfo{person}{Jiacheng Li},
  {and} \bibinfo{person}{Julian~J. McAuley}.} \bibinfo{year}{2019}\natexlab{}.
\newblock \showarticletitle{Justifying Recommendations using Distantly-Labeled
  Reviews and Fine-Grained Aspects}. In
  \bibinfo{booktitle}{\emph{EMNLP-IJCNLP}}. \bibinfo{publisher}{Association for
  Computational Linguistics}.
\newblock


\bibitem[\protect\citeauthoryear{Ravi}{Ravi}{2019}]%
        {DBLP:conf/icml/Ravi19}
\bibfield{author}{\bibinfo{person}{Sujith Ravi}.}
  \bibinfo{year}{2019}\natexlab{}.
\newblock \showarticletitle{Efficient On-Device Models using Neural
  Projections}. In \bibinfo{booktitle}{\emph{ICML}}.
\newblock


\bibitem[\protect\citeauthoryear{Rendle, Freudenthaler, Gantner, and
  Schmidt{-}Thieme}{Rendle et~al\mbox{.}}{2009}]%
        {DBLP:conf/uai/RendleFGS09}
\bibfield{author}{\bibinfo{person}{Steffen Rendle}, \bibinfo{person}{Christoph
  Freudenthaler}, \bibinfo{person}{Zeno Gantner}, {and} \bibinfo{person}{Lars
  Schmidt{-}Thieme}.} \bibinfo{year}{2009}\natexlab{}.
\newblock \showarticletitle{{BPR:} Bayesian Personalized Ranking from Implicit
  Feedback}. In \bibinfo{booktitle}{\emph{UAI}}.
\newblock


\bibitem[\protect\citeauthoryear{Sennrich, Haddow, and Birch}{Sennrich
  et~al\mbox{.}}{2016}]%
        {DBLP:conf/acl/SennrichHB16a}
\bibfield{author}{\bibinfo{person}{Rico Sennrich}, \bibinfo{person}{Barry
  Haddow}, {and} \bibinfo{person}{Alexandra Birch}.}
  \bibinfo{year}{2016}\natexlab{}.
\newblock \showarticletitle{Neural Machine Translation of Rare Words with
  Subword Units}. In \bibinfo{booktitle}{\emph{ACL}}.
\newblock


\bibitem[\protect\citeauthoryear{Serr{\`{a}} and Karatzoglou}{Serr{\`{a}} and
  Karatzoglou}{2017}]%
        {bloom}
\bibfield{author}{\bibinfo{person}{Joan Serr{\`{a}}} {and}
  \bibinfo{person}{Alexandros Karatzoglou}.} \bibinfo{year}{2017}\natexlab{}.
\newblock \showarticletitle{Getting Deep Recommenders Fit: Bloom Embeddings for
  Sparse Binary Input/Output Networks}. In \bibinfo{booktitle}{\emph{RecSys}}.
  \bibinfo{publisher}{{ACM}}.
\newblock


\bibitem[\protect\citeauthoryear{Shannon}{Shannon}{1948}]%
        {DBLP:journals/bstj/Shannon48}
\bibfield{author}{\bibinfo{person}{Claude~E. Shannon}.}
  \bibinfo{year}{1948}\natexlab{}.
\newblock \showarticletitle{A mathematical theory of communication}.
\newblock \bibinfo{journal}{\emph{Bell Syst. Tech. J.}} \bibinfo{volume}{27},
  \bibinfo{number}{3} (\bibinfo{year}{1948}), \bibinfo{pages}{379--423}.
\newblock


\bibitem[\protect\citeauthoryear{Shi, Mudigere, Naumov, and Yang}{Shi
  et~al\mbox{.}}{2020}]%
        {fb}
\bibfield{author}{\bibinfo{person}{Hao{-}Jun~Michael Shi},
  \bibinfo{person}{Dheevatsa Mudigere}, \bibinfo{person}{Maxim Naumov}, {and}
  \bibinfo{person}{Jiyan Yang}.} \bibinfo{year}{2020}\natexlab{}.
\newblock \showarticletitle{Compositional Embeddings Using Complementary
  Partitions for Memory-Efficient Recommendation Systems}. In
  \bibinfo{booktitle}{\emph{SIGKDD}}.
\newblock


\bibitem[\protect\citeauthoryear{Sitzmann, Martel, Bergman, Lindell, and
  Wetzstein}{Sitzmann et~al\mbox{.}}{2020}]%
        {siren}
\bibfield{author}{\bibinfo{person}{Vincent Sitzmann}, \bibinfo{person}{Julien
  N.~P. Martel}, \bibinfo{person}{Alexander~W. Bergman},
  \bibinfo{person}{David~B. Lindell}, {and} \bibinfo{person}{Gordon
  Wetzstein}.} \bibinfo{year}{2020}\natexlab{}.
\newblock \showarticletitle{Implicit Neural Representations with Periodic
  Activation Functions}.
\newblock \bibinfo{journal}{\emph{CoRR}}  \bibinfo{volume}{abs/2006.09661}
  (\bibinfo{year}{2020}).
\newblock
\showeprint[arxiv]{2006.09661}


\bibitem[\protect\citeauthoryear{Svenstrup, Hansen, and Winther}{Svenstrup
  et~al\mbox{.}}{2017}]%
        {hashemb}
\bibfield{author}{\bibinfo{person}{Dan Svenstrup},
  \bibinfo{person}{Jonas~Meinertz Hansen}, {and} \bibinfo{person}{Ole
  Winther}.} \bibinfo{year}{2017}\natexlab{}.
\newblock \showarticletitle{Hash Embeddings for Efficient Word
  Representations}. In \bibinfo{booktitle}{\emph{NIPS}}.
\newblock


\bibitem[\protect\citeauthoryear{Tancik, Srinivasan, Mildenhall,
  Fridovich{-}Keil, Raghavan, Singhal, Ramamoorthi, Barron, and Ng}{Tancik
  et~al\mbox{.}}{2020}]%
        {fourier}
\bibfield{author}{\bibinfo{person}{Matthew Tancik}, \bibinfo{person}{Pratul~P.
  Srinivasan}, \bibinfo{person}{Ben Mildenhall}, \bibinfo{person}{Sara
  Fridovich{-}Keil}, \bibinfo{person}{Nithin Raghavan},
  \bibinfo{person}{Utkarsh Singhal}, \bibinfo{person}{Ravi Ramamoorthi},
  \bibinfo{person}{Jonathan~T. Barron}, {and} \bibinfo{person}{Ren Ng}.}
  \bibinfo{year}{2020}\natexlab{}.
\newblock \showarticletitle{Fourier Features Let Networks Learn High Frequency
  Functions in Low Dimensional Domains}. In
  \bibinfo{booktitle}{\emph{NeurIPS}}.
\newblock


\bibitem[\protect\citeauthoryear{Weinberger, Dasgupta, Langford, Smola, and
  Attenberg}{Weinberger et~al\mbox{.}}{2009}]%
        {hashing}
\bibfield{author}{\bibinfo{person}{Kilian~Q. Weinberger},
  \bibinfo{person}{Anirban Dasgupta}, \bibinfo{person}{John Langford},
  \bibinfo{person}{Alexander~J. Smola}, {and} \bibinfo{person}{Josh
  Attenberg}.} \bibinfo{year}{2009}\natexlab{}.
\newblock \showarticletitle{Feature hashing for large scale multitask
  learning}. In \bibinfo{booktitle}{\emph{ICML}}.
\newblock


\bibitem[\protect\citeauthoryear{Yi, Yang, Hong, Cheng, Heldt, Kumthekar, Zhao,
  Wei, and Chi}{Yi et~al\mbox{.}}{2019}]%
        {DBLP:conf/recsys/YiYHCHKZWC19}
\bibfield{author}{\bibinfo{person}{Xinyang Yi}, \bibinfo{person}{Ji Yang},
  \bibinfo{person}{Lichan Hong}, \bibinfo{person}{Derek~Zhiyuan Cheng},
  \bibinfo{person}{Lukasz Heldt}, \bibinfo{person}{Aditee Kumthekar},
  \bibinfo{person}{Zhe Zhao}, \bibinfo{person}{Li Wei}, {and}
  \bibinfo{person}{Ed~H. Chi}.} \bibinfo{year}{2019}\natexlab{}.
\newblock \showarticletitle{Sampling-bias-corrected neural modeling for large
  corpus item recommendations}. In \bibinfo{booktitle}{\emph{RecSys}}.
  \bibinfo{publisher}{{ACM}}.
\newblock


\bibitem[\protect\citeauthoryear{Zhang, Liu, Xie, Ktena, Tejani, Gupta, Myana,
  Dilipkumar, Paul, Ihara, Upadhyaya, Huszar, and Shi}{Zhang
  et~al\mbox{.}}{2020}]%
        {twitter}
\bibfield{author}{\bibinfo{person}{Caojin Zhang}, \bibinfo{person}{Yicun Liu},
  \bibinfo{person}{Yuanpu Xie}, \bibinfo{person}{Sofia~Ira Ktena},
  \bibinfo{person}{Alykhan Tejani}, \bibinfo{person}{Akshay Gupta},
  \bibinfo{person}{Pranay~Kumar Myana}, \bibinfo{person}{Deepak Dilipkumar},
  \bibinfo{person}{Suvadip Paul}, \bibinfo{person}{Ikuhiro Ihara},
  \bibinfo{person}{Prasang Upadhyaya}, \bibinfo{person}{Ferenc Huszar}, {and}
  \bibinfo{person}{Wenzhe Shi}.} \bibinfo{year}{2020}\natexlab{}.
\newblock \showarticletitle{Model Size Reduction Using Frequency Based Double
  Hashing for Recommender Systems}. In \bibinfo{booktitle}{\emph{RecSys}}.
\newblock


\end{thebibliography}

\newpage
\appendix

\noindent\textbf{\huge{Appendix}}

\vspace{0.2cm}

\noindent\textbf{\Large{A. Analysis on Encoding Properties}}

\vspace{0.1cm}

We formally define and analyze the encoding properties. For demonstration, we use a setting similar to what we used in the experiments: $n$=$10^6$, $m$=$10^6$, and $k=1024$.

\vspace{0.1cm}
\noindent\textbf{\large{A.1 Uniqueness}}
\vspace{0.1cm}

\theoremstyle{definition}
\begin{definition}[Uniqueness in encoding]
An encoding function $E$ is a unique encoding if $P(E(x))=P(E(y))<\epsilon, \forall x, y\in V$, where $\epsilon$ is a near-zero constant.
\end{definition}
Obviously the \emph{identity encoding}, \emph{one-hot encoding}, and \emph{binary encoding} satisfy the uniqueness property.

For hashing methods, the probability of having collision is $1-e^{-\frac{n(n-1)}{2m}}$ where $m$ is the total number of hashing buckets ($m^2$ buckets for double hashing), according to~\cite{DBLP:conf/stoc/CarterW77}. The probability is 1.0, and 0.39 for \emph{one-hot hashing} and \emph{double one-hot hashing}, respectively. For DHE, the number of possible hashing buckets is $m^k=10^{6144}$, and the collision rate is extremely small. Thus we can safely assume there is no collision.

\vspace{0.1cm}
\noindent\textbf{\large{A.2 Equal Similarity}}
\vspace{0.1cm}

\begin{definition}[Equal similarity in encoding]
An encoding functions $E$ is a equally similar encoding if $\mathbb{E}[\text{Euclidean\_distance}(E(x)-E(y))]=c, \forall x, y\in V$, where $c$ is a non-zero constant.
\end{definition}

Obviously the \emph{identity encoding}, and \emph{binary encoding} don't satisfy the property.

For one-hot based hashing methods, the expected Euclidean distance is $k*\frac{m-1}{m}$. For DHE, the expectation is:
\begin{equation*}
    \begin{split}
        \mathbb{E}[(E(x)-E(y))^2]=& \mathbb{E}[E^2(x) -2E(x)E(y)+E^2(y)]\\
        =&\mathbb{E}[E^2(x)]-2\mathbb{E}[E(x)]\mathbb{E}[E(y)]+\mathbb{E}[E^2(y)]\\
        =&\frac{m(2m+1)(m+1)}{3}-\frac{(m+1)^2}{2}
    \end{split}
\end{equation*}

\vspace{0.1cm}
\noindent\textbf{\large{A.3 High Dimensionality}}
\vspace{0.1cm}

This is a subjective property, and we generally think larger than 100-dim can be considered as high-dimensional spaces. Following this, the 1-dim \emph{identity encoding} and the $\left\lceil\log n\right\rceil$-dim \emph{binary encoding} doesn't satisfy the property.

\vspace{0.1cm}
\noindent\textbf{\large{A.4 High Shannon Entropy}}
\vspace{0.1cm}

\begin{definition}[High Shannon Entropy]
An encoding functions $E$ has the high entropy property if for any dimension $i$, the entropy $H(E(x)_i)=H^*, (x\in V)$, where $H^*=\log o$ is the max entropy for $o$ outcomes (e.g. $H^*=1$ for binary outcome).
\end{definition}

As the zeros and ones are uniformly distributed at each dimension in \emph{binary encoding}, the entropy equals to $H^*=1$. Similarly, the entropy of \emph{identity encoding} also reaches the maximal entropy $H=-\sum_{i=1}^{n}\frac{1}{n}\log\frac{1}{n}=\log n=H^*$.

For one-hot full embedding, at each dimension, the probability is $\frac{1}{n}$ for having 1, and $\frac{n-1}{n}$ for having 0. So the entropy $H=-\frac{1}{n}\log\frac{1}{n}-\frac{n-1}{n}\log\frac{n-1}{n}$, which quickly converges to zero with a large $n$. The entropy is significantly less than $H^*=1$

For one-hot hashing, the probability of having ones is $\frac{n}{m}\cdot\frac{1}{n}=\frac{1}{m}$, and for zeros it's $\frac{m-1}{m}$. Therefore the entropy $H=-\frac{1}{m}\log\frac{1}{m}-\frac{m-1}{m}\log\frac{m-1}{m}$, which is near zero due to the large $m$. Double one-hot hashing has a similar conclusion.

For DHE, at each dimension, the encodings are uniformly distributed among $[m]=\{1,2,\dots,m\}$. Therefor the entropy $H=-\sum_{i=1}^{m}\frac{1}{m}\log\frac{1}{m}=\log m=H^*$, which reaches the maximal entropy.

\begin{table}[t]
\small
\caption{Notation. \label{tb:notation}}\vspace{-0.3cm}
\begin{tabularx}{\linewidth}{lX}
\toprule
Notation&Description\\
\midrule
$V$                & set of feature values\\
$n\in \mathbb{N}$  & vocabulary size\\
$d\in \mathbb{N}$  & embedding dimension\\
$m\in \mathbb{N}$  & hashed vocabulary size (usually $m<n$)\\
$H:V\to [m]$                & hash function mapping feature values to $\{1,2,\dots,m\}$\\
$k\in \mathbb{N}$  & number of hash functions, also the encoding length in DHE\\
$d_{\text{NN}}\in \mathbb{N}$  & the width of hidden layers in the embedding network\\
$h\in \mathbb{N}$  & the number of hidden layers in the embedding network\\
\bottomrule
\end{tabularx}

\end{table}

\vspace{0.2cm}

\noindent\textbf{\Large{B. Experimental Setup}}

\vspace{0.1cm}
\noindent\textbf{\large{B.1 Dataset Processing}}
\vspace{0.1cm}

\begin{algorithm}[t]
\caption{Deep Hash Embedding (DHE).}\label{algo:dhe}
\small
\DontPrintSemicolon
  
  \KwInput{a feature value $x\in\mathbb{N}$, encoding length $k$, embedding dim $d$, memory budget $B$, network depth $h$}
  \KwOutput{$emb\in\mathbb{R}^{d}$, a $d$-dim embedding for $x$}
  
  \tcc{Calculate the dense hash encoding (parameter-free)}
  $encod\leftarrow \text{DenseHashEncoding(x)}$
  
  \tcc{Define the learnable variables in DNN}
  $F \leftarrow \text{BuildingDNN}(k, d, h, B)$
  
  \tcc{Feed the encoding vector into DNN for generating the embedding}
  $emb \leftarrow F(encod)$
\end{algorithm}

We use two commonly used public benchmark datasets for evaluating recommendation performance:
\begin{itemize}[leftmargin=4mm]
\item{\textbf{Movielens-20M}} is a widely used benchmark for evaluating collaborative filtering algorithms~\cite{DBLP:journals/tiis/HarperK16}. The dataset includes 20M user ratings on movies\footnote{\url{https://grouplens.org/datasets/movielens/20m/}}.
\item{\textbf{Amazon Books}} is the largest category in a series of datasets introduced in~\cite{DBLP:conf/emnlp/NiLM19}, comprising large corpora of product reviews crawled from \emph{Amazon.com}. We used the latest 5-core version crawled in 2018\footnote{\url{https://nijianmo.github.io/amazon/index.html}}. The dataset is known for its high sparsity.
\end{itemize}

\begin{table}[t]
\centering
\caption{Dataset statistics\label{tab:data}}\vspace{-0.3cm}
\small
\begin{tabular}{lrrrrr}
\toprule
 Dataset 				&  \#users & \#items & total vocab size &  \#actions & sparsity\\ \midrule
\emph{MovieLens}     		&  138K  &  27K & 165K	&	20M&99.47\%\\
\emph{Amazon}    	    &  1.9M	 & 0.7M  & 2.6M	&	27M &99.99\%\\ \bottomrule
\end{tabular}
\end{table}

The dataset statistics are shown in Table~\ref{tab:data}. As in~\cite{DBLP:conf/recsys/HeKM17}, we treat all ratings as observed feedback, and sort the feedback according to timestamps. For each user, we withhold their last two actions, and put them into the validation set and test set respectively. All the rest are used for model training.

\vspace{0.1cm}
\noindent\textbf{\large{B.2 Implementation Details \& Hyper-parameters}}
\vspace{0.1cm}

We implement all the methods using \emph{TensorFlow}. The embedding dimension $d$ for user and item embeddings is set to 32 for the best Full Emb performance, searched among $\{8, 16, 32, 64\}$, for both datasets. For the recommendation model training, we use the Adam optimizer with a learning rate of 0.001. We apply the embedding schemes on both user and item embeddings. The initialization strategy follows~\cite{DBLP:conf/www/HeLZNHC17}. The model training is accelerated with a single NVIDIA V-100 GPU. To reduce the variance, all the results are the average of the outcomes from 5 experiments.

For HashEmb~\cite{hashemb}, we use dedicated weights (without collision) for each feature value for better performance. For Hybrid Hashing~\cite{twitter}, we use dedicated embeddings for the top 10\% of the most frequent feature values, and apply double hashing for the others. By default, we use $k$=2 hash functions for hashing-based baselines (except for the hashing trick~\cite{hashing} which uses a single hash function), which is suggested by the authors~\cite{hashemb, twitter, bloom}. The given model size budget decides the hashed vocabulary size $m$ for hashing-based methods (e.g. a half of the full model size means $m$=$n/2$). For compositional embedding~\cite{fb}, we use the quotient-remainder trick to generate two complementary hashing, and adopt the path-based variant with a MLP with one hidden layer of 64 nodes as used in the paper.

For DHE we use the same hyper-parameters for both datasets: $k$=1024 hash functions to generate the hash encoding vector, followed by a 5-layer feedforward neural network with Batch Normalization~\cite{bn} and Mish activation function~\cite{mish}. The width $d_{\text{NN}}$ of the network is determined by the given model size. The $m$ in DHE is set to $10^6$.

\vspace{0.2cm}

\noindent\textbf{\Large{C. Pseudo-code}}

\vspace{0.1cm}

Algorithm~\ref{algo:dhe} presents the overall process of the proposed Deep Hash Embedding~(DHE). Algorithm~\ref{algo:encod} presents the encoding process of the dense hash encoding. Algorithm~\ref{algo:transform} presents the transformations for converting the integer vector (after hashing) into real-valued vectors and approximating a uniform or Gaussian distribution. We utilize the evenly distributed property of universal hashing~\cite{DBLP:conf/stoc/CarterW77} to build the uniform distribution, and adopts the Box-Muller transform\footnote{\url{https://en.wikipedia.org/wiki/Box\%E2\%80\%93Muller_transform}} to construct a Gaussian distribution from pairs of uniformly distributed samples ($U(0,1)$)~\cite{box1958note}.

\begin{algorithm}[htp]
\caption{Dense Hash Encoding in DHE (on-the-fly).}\label{algo:encod}
\small
\DontPrintSemicolon
  
  \KwInput{a feature value $x\in\mathbb{N}$, encoding length $k$, hash buckets $m$}
  \KwOutput{$encod\in\mathbb{R}^{k}$, a $k$-dim dense hash encoding for $x$}
  \tcc{Using a fixed seed to generate the same hash functions at each encoding process. The generation can be skipped via storing $O(k)$ parameters for hashing.}
  Set the Random Seed to 0.
  
  \For{$i\leftarrow 1$ \KwTo $k$}
  {
  \tcc{$a$ and $b$ are randomly chosen integer with $b\neq 0$, $p$ is a prime larger than $m$}
  
  $a \leftarrow \text{RandomInteger()}$
  
  $b \leftarrow \text{RandomNonZeroInteger()}$
  
  $p \leftarrow \text{RandomPrimerLargerThan}(m)$
  
  \tcc{Applies universal hashing for integers}
  
  $h[i]\leftarrow ((ax+b)\ \text{mod}\ p)\ \text{mod}\ m$
  }
  
  \tcc{Apply a transformation to get the real-valued encoding}
  $encod \leftarrow \text{Transform}(h)$
\end{algorithm}

\begin{algorithm}[htp]
\caption{Encoding Transform.}\label{algo:transform}
\small
\DontPrintSemicolon
  
  \KwInput{$h\in\{1,2,\dots,m\}^{k}$, $k$ indices of hashing buckets.}
  \KwOutput{$encod\in\mathbb{R}^{k}$, a $k$-dim dense hash encoding for $x$}
  
  \For{$i\leftarrow 1$ \KwTo $k$}
  {
    $encod'[i] \leftarrow (h[i] - 1)/(m-1)$\tcp*{$encod'[i]\in[0, 1]$}
    
    $encod[i] \leftarrow encod'[i] * 2 -1$\tcp*{$encod[i]\in[-1, 1]$}
  }
  
  \eIf{the distribution is uniform}{
    \Return $encod$\tcp*{uniform distribution $U(-1,1)$}
  }{
    
    \tcc{Box-Muller Transform for Gaussian distribution}
    $i\leftarrow 0$
    
    \While{$i<m$}
    {
      $j\leftarrow i + 1$
      
      $encod[i] \leftarrow \sqrt{-2\ln encod'[i]}\cos(2\pi encod'[j])$ 
      
      $encod[j] \leftarrow \sqrt{-2\ln encod'[i]}\sin(2\pi encod'[j])$
      
      $i\leftarrow i + 2$
    }
    
    \Return $encod$\tcp*{Gaussian distribution $\mathcal{N}(0,1)$}
  }
  
\end{algorithm}

\end{document}